\documentclass[bimj,fleqn]{w-art}
\usepackage{times}
\usepackage{w-thm}
\usepackage[authoryear]{natbib}

\usepackage{bm}
\usepackage{float}
\usepackage{subfloat}
\usepackage{tikz, pict2e}
\usepackage{graphicx}
\usepackage[caption=false]{subfig}
\usepackage{makecell}
\usepackage{amsthm,amsmath,amsfonts,amssymb}
\usepackage[colorlinks,citecolor=blue,urlcolor=blue]{hyperref}

\theoremstyle{plain}

\theoremstyle{definition}

\usepackage[]{graphicx}
\chardef\bslash=`\\ 

\makeatletter
\DeclareRobustCommand{\dplus}{\DOTSB\mathop{\dplus@}\slimits@}
\newcommand{\dplus@}{\vphantom{\sum}\mathpalette\dplus@@\relax}
\newcommand{\dplus@@}[2]{%
  \begingroup
  \sbox\z@{$#1\sum$}%
  \unitlength=\dimexpr\ht\z@+\dp\z@\relax
  \linethickness{%
    \ifx#1\displaystyle 1.8\fontdimen8\textfont3 \else
    \ifx#1\textstyle 1.2\fontdimen8\textfont3 \else
    \ifx#1\scriptstyle 1.2\fontdimen8\scriptfont3 \else
    1.3\fontdimen8\scriptscriptfont3 \fi\fi\fi}
  \vcenter{\hbox{%
    \begin{picture}(1,1)
    \polygon(0.05,0.5)(0.5,0.05)(0.95,0.5)(0.5,0.95)
    \Line(0.5,0.05)(0.5,0.95)
    \Line(0.05,0.5)(0.95,0.5)
    \end{picture}%
  }}%
  \endgroup
}
\makeatother

\begin{document}
\keywords{BAMMIT MODEL; BAYESIAN INFERENCE; TENSORS\\
}  

\title[An Extension to Bayesian AMMI Models Using Tensor Regression]{Bayesian Additive Main Effects and Multiplicative Interaction Models Using Tensor Regression for Multi-environmental Trials}
\author[dos Santos {\it{et al.}}]{Antonia A. L. dos Santos\footnote{Corresponding author: {\sf{e-mail: alessalemos23@gmail.com}}}\inst{,1}} 
\address[\inst{1}]{Hamilton Institute, Department of Mathematics and Statistics, Maynooth University, Ireland}
\author[]{Andrew C. Parnell\inst{2}}
\address[\inst{2}]{School of Mathematics and Statistics, Insight Centre for Data Analytics, University College Dublin, Ireland}
\author[]{Rafael A. Moral\inst{1}} 
\author[]{Danilo A. Sarti\inst{1,2}}

\begin{abstract}
We propose a Bayesian tensor regression model to accommodate the effect of multiple factors on phenotype prediction. We adopt a set of prior distributions that resolve identifiability issues that may arise between the parameters in the model. Further, we incorporate a spike-and-slab structure that identifies which interactions are relevant for inclusion in the linear predictor, even when they form a subset of the available variables. Simulation experiments show that our method outperforms previous related models and machine learning algorithms under different sample sizes and degrees of complexity. We further explore the applicability of our model by analysing real-world data related to wheat production across Ireland from 2010 to 2019. Our model performs competitively and overcomes key limitations found in other analogous approaches. Finally, we adapt a set of visualisations for the posterior distribution of the tensor effects that facilitate the identification of optimal interactions between the tensor variables, whilst accounting for the uncertainty in the posterior distribution.
\end{abstract}

\maketitle                   






\section{Introduction}

The phenotypic performance of a cultivar is associated with many potentially interacting variables \citep{hara2021selection}. These may include, but are not limited to: genetic factors, environmental exposure, soil type, climatic conditions, and season. Any combinations of these factors may contribute, either positively or negatively, to the variability of the production of the crop of interest \citep{kross2020using}. Statistical modelling of the effect of these variables, both singly and jointly, is an important decision-making tool for farmers and those in the agricultural sector for predicting e.g. yield \citep{adisa2019application}.

One of the main interactions believed to impact most on crop production is between genotype and environment, which we denote for notational convenience as $G\times E$. This sort of interaction is characterised by cultivars that do not behave consistently in differing environments. Therefore, it is necessary to estimate the amount of variation in crop yield that is caused by it. Many models have been proposed to estimate $G\times E$ \citep{gauch2008statistical, crossa2010linear, gauch2013simple}. The most popular is the additive main effects and multiplicative interaction (AMMI) model \citep{gauch1988model}, which consists of two components. The first term is the additive component, which contains the main effects of categorically structured genotype and environmental factors. The second term involves a sum of a multiplication of parameters, which are constrained to an orthonormal space and represent how strong/weak the interactions between the genotypes and environments are. The AMMI model is restricted to using only these two covariates. While genotype and environment interactions are key components in predicting crop yield, a comprehensive understanding requires the consideration of additional factors.  

Our approach allows for more components beyond genotype and environment to be included in the AMMI model. We follow the Bayesian tensor regression technique of \citet{guhaniyogi2017bayesian} to allow for any number of interacting categorical factors. Tensors are algebraic structures that generalise matrices and provide a generic way of describing multidimensional arrays on a given number of axes. Tensor decomposition methods have the advantage of capturing the information in the data with a multi-linear structure  and bring a unique representation without the requirement for additional constraints like sparsity or statistical independence \citep{jorgensen2018probabilistic}. The two main tensor decompositions are the PARAFAC \citep{carroll1970analysis, harshman1970foundations} and Tucker models \citep{tucker1963implications}. Tensors have been used in many fields of study, including physics \citep{gaillac2016elate}, chemistry \citep{facelli2011chemical}, medicine \citep{peyrat2007computational}, and data mining \citep{morup2011applications}. \cite{guhaniyogi2017bayesian} propose a tensor-based Bayesian regression model where vector/tensor covariates are used to estimate a univariate response through a class of multiway shrinkage priors. They illustrate the model on real-world data from the brain connectome and provide theoretical results concerning the speed at which the posterior distribution converges to the true posterior (i.e., the contraction rate). Similarly, \cite{papadogeorgou2021soft} propose a soft tensor regression to investigate the connection between human traits and brain structural connectomics. 

In this paper, we propose the Bayesian Additive Main effects and Multiplicative Interaction Tensor (BAMMIT) model, which generalises the AMMI model to contain a tensor of interacting terms. We extend the standard AMMI model to include new parameters to the additive and multiplicative terms of the  model, taking into account factors other than genotype and environment on the phenotype of a given cultivar. Common extra factors might include soil types, time points, or growth stages. Our methodology captures both minor and major interactions of these factors and identifies which of these are important and need to be included in the model. We present our new model in a Bayesian hierarchical format where we place prior distributions on the main and tensor product terms so as to guarantee the model's identifiability  and impose orthonormality constraints, which are an essential part of both the original AMMI and the BAMMIT models. 

We evaluate our proposed approach through a set of simulation experiments.  Our interest is to investigate the model's performance under different sample sizes and degrees and complexity of interaction structures. We compare the predictions from our model with other machine learning models and standard linear mixed effects models in terms of the out-of-sample root mean squared error (RMSE) and the coefficient of determination ($R^2$).

The remainder of this paper is structured as follows. In Section \ref{secammi}, we review the AMMI model and present the constraints imposed on its two components. In Section \ref{secbammit}, we introduce our BAMMIT model with its extended additive and multiplicative terms. We outline the interpretability and identifiability constraints, as well as the prior distributions considered for the parameters and a description of how we obtain the posterior distributions. In Section \ref{simulation}, we compare the results from BAMMIT with other relevant models based on synthetic data. In Section \ref{case}, we analyse real-world data on wheat production in Ireland from 2010 to 2019. We also utilise a new set of visualisations to assess the posterior distributions of the components of the BAMMIT model, and identify optimal interactions and their associated uncertainty. Finally, we review and discuss the findings of the work in Section \ref{discussion}. 

\section{Methods} \label{sec:model}

In this section, we review the original AMMI model and define terminology and notation. We then introduce the BAMMIT model detailing the necessary constraints to ensure identifiability as well as the prior distributions and inferential scheme. In real-world scenarios the models we introduce below often include additional replicate or block effects that complicate the hierarchical structure of the data. However, we remove these for simplicity of exposition in our model definitions, and include them only in our case studies.

\subsection{AMMI model}\label{secammi}

The traditional AMMI model takes into account only two categorical factors, commonly genotype and environment, and is given by a combination of two parts, one additive and one multiplicative. Let $y_{ij}$ be the outcome variable with $i$ representing genotype and $j$ environment. We write the model as:
\begin{eqnarray}\label{eq:ammi}
y_{ij}=\mu+b_{i}^{(1)}+b_{j}^{(2)}+\sum_{q=1}^{Q} \lambda_{q}\beta_{i q}^{(1)} \beta_{j q}^{(2)}+\varepsilon_{ij}, \mbox{ } \varepsilon_{ij} \sim \mbox{N}(0, \sigma^{2}),
\end{eqnarray}
\noindent
where $b_{i}^{(1)}$ and $b_{j}^{(2)}$ represent the marginal effect of the $i^{th}$ genotype and $j^{th}$ environment, respectively, $i = 1,...,B_1$ and $j = 1,..., B_2$. The bilinear term (i.e. the summation) is composed of $Q$ components, each of which contains a variable $\lambda_{q}$ and the scores $\beta_{i q}^{ (1)}$ and $\beta_{j q}^{(2)}$. The parameter $\lambda_{q}$ measures the interaction strength of the $q^{th}$ component and is usually ordered such that $\lambda_{1}\geq\lambda_{2}\geq\dots\geq\lambda_{Q}$. The scores $\beta_{i q}^{ (1)}$ and $\beta_{j q}^{(2)}$  represent the importance of the $i^{th}$ genotype and the $j^{th}$ environment in the interaction. To ensure identifiability, the bilinear term is constrained so that $\sum_{i} \beta_{iq}^{(1)} \beta_{i q'}^{(1)} = \sum_{j} \beta_{jq}^{(2)} \beta_{j q'}^{(2)} = 0$, for $q \neq q'$ and $\sum_{i} (\beta_{iq}^{(1)})^2  = \sum_{j} (\beta_{jq}^{(2)})^2 = 1$. All the terms in Equation \ref{eq:ammi} are estimated with the exception of $Q$ which is fixed.

There are a range of approaches for estimating the parameters of the AMMI model. In the frequentist paradigm, the additive term of Equation (\ref{eq:ammi}) is estimated by  ordinary least squares ignoring the interaction term, and subsequently a singular value decomposition (SVD) of the matrix of residuals is used to estimate the multiplicative terms \citep{gabriel1978least}. In this case, in addition to the constraints applied to the multiplicative term, it is generally ensured that $\sum_ib^{(1)}_i = \sum_jb^{(2)}_j = 0$, in order to maintain orthogonality between the main effects and interaction terms, as well as to directly interpret the parameters.

Within the Bayesian context,  \cite{viele2000parsimonious} proposed the use of Markov chain Monte Carlo (MCMC) to estimate the parameters of the AMMI model ensuring that the inherent constraints of the model are not violated. \cite{liu2001bayesian} formulated a more stable and computationally faster Gibbs sampler. \cite{crossa2011bayesian} and \cite{perez2012general} proposed a Gibbs sampler such that the algorithm was stabilised and incorporated statistical inference in the visualisation of biplots \citep{gabriel1971biplot}, drawing credibility regions for the interaction effects. By contrast, \cite{josse2014another} introduced an approach to deal with the overparametrization issue of the model by defining priors for the complete set of parameters ignoring the constraints, then applying a postprocessing on the posterior samples of each parameter. \cite{sarti2023bayesian} used Bayesian additive regression trees (BART) in which a `double-grow' BART is responsible for capturing the interaction term.

As stated above, the number of terms in the summation, $Q$, is usually fixed. It is assumed that $Q\leq \mbox{min}(B_1 - 1, B_2 - 1)$. The total variability measured by the principal components is linked to $Q$, such that by setting $Q =  \mbox{min}(B_1 - 1, B_2 - 1)$ the model can capture all the variance in the interaction. In practice, Q is commonly an integer between 1 and 3 as this allows for easier interpretation and visualisation of the interaction effects via biplots. However, many approaches can be applied to determine the value of $Q$. Examples include \cite{cornelius1993statistical} who applied parametric significance tests; other authors who employed cross validation techniques \citep{dos2003model, gabriel2002biplot, hadasch2017cross}, or those using resampling techniques \citep{malik2018nonparametric, malik2019testing}. Examples in the Bayesian field include \cite{perez2012general} and \cite{da2015bayesian} where the prior choice and Bayes factor deal with determining the number of components of the multiplicative term. The non-parametric Bayesian approach of \cite{sarti2023bayesian} bypasses the need to provide $Q$ completely but, like many BART models, suffers from interpretability problems due to the complexity of the regression trees. 

One of the reasons for the popularity of the AMMI model is its strong predictive performance \citep{gauch2006statistical, gauch2008statistical}, accuracy \citep{gauch2019ammisoft} and its stability evaluation system \citep{gauch1988model,yue2022genotype}. Given its desirable properties, many extensions can be found in the literature, as highlighted above. In this work, we aim to maintain the structure of the AMMI model and add the effects of other categorical factors that are commonly available in real-world multi-environmental trials.


\subsection{BAMMIT model}\label{secbammit}

The model in (\ref{eq:ammi}) can be extended to include the effect of many factors apart from genotype and environment. Let $y_{i j \ldots v}$ be an outcome variable, in a setting with a total of $N$ observations and $V$ predictors. We define the BAMMIT model as:
\begin{eqnarray}\label{bammit0}
y_{i j \ldots v}=\mu+b_{i}^{(1)}+b_{j}^{(2)}+\cdots+b_{v}^{(V)}+\sum_{q=1}^{Q} \lambda_{q}\left(\beta_{i q}^{(1)} \beta_{j q}^{(2)} \times \cdots \times \beta_{v q}^{(V)}\right)+\varepsilon_{i j \ldots v},
\end{eqnarray}
where $\varepsilon_{i j \ldots v} \sim \mbox{N}(0, \sigma^{2})$. This is similar to the AMMI model described in (\ref{eq:ammi}), however now we have $V$ factors instead of only two. Alternatively, we can rewrite the coefficients of the additive and multiplicative terms of (\ref{bammit0}) in tensor notation. Let  $\bm{b}^{(v)}=(b_1^{(v)}, \dots, b_{B_v}^{(v)})^\top$ be a $B_v$-dimensional vector of parameters of the $v^{th}$ predictor and $\bm{\beta}_q^{(v)} = (\beta_{1q}^{(v)}, \dots, \beta_{B_vq}^{(v)})^\top$ be a $B_v$-dimensional vector of singular values, with $q = 1,..., Q$. Binding the column vectors $\bm{\beta}_q^{(v)}$, we get $\bm{\beta}^{(v)}$, a matrix of dimension $B_v \times Q$. We define $N = \left(\prod_{v=1}^V B_v\right)$ as the total number of observations (though, for example, replication may increase $N$ without any need for extra parameters). 

For notational convenience, we define a cumulative direct sum and a cumulative Kronecker product resulting in an $N$-dimensional vector as $\dplus_{v=1}^V \mathbf{b}^{(v)}=\mathbf{b}^{(1)} {\scriptstyle\dplus} \cdots {\scriptstyle\dplus} \text{\hspace{0.1cm}} \bm{b}^{(V)}$  and $\bigotimes_{v=1}^{V} \boldsymbol{\beta}_{q}^{(v)}=\boldsymbol{\beta}_{q}^{(1)} \otimes \cdots \otimes \boldsymbol{\beta}_{q}^{(V)}$, respectively. The direct sum operation is defined such that for vectors $\mathbf{a}=\left(a_{1}, a_{2}\right)^{\top}$  and  $\mathbf{b}=\left(b_{1}, b_{2}, b_{3}\right)^{\top}$, for example, $\mathbf{a} \text{\hspace{0.1cm}}{\scriptstyle\dplus} \text{\hspace{0.1cm}} \mathbf{b}=\left(a_{1}+b_{1}, a_{1}+b_{2}, a_{1}+b_{3}, a_{2}+b_{1}, a_{2}+b_{2}, a_{2}+b_{3}\right)^{\top}.$

Following the tensor notation presented, the BAMMIT model can be written more compactly as:
\begin{eqnarray}\label{bmodel}
\bm{y} &=& \mu\bm{1}^\top_N + \dplus_{v=1}^V \bm{b}^{(v)} + \sum\limits_{q=1}^Q \lambda_q\left( \bigotimes_{v = 1}^{V} \bm{\beta}^{(v)}_q\right)+\bm{\varepsilon},
\end{eqnarray}

\noindent
where $\bm{y}$ is an $N$-dimensional vector, as before $\mu$ is the grand mean, $\lambda_q$ is the strength of the $q^{th}$ component, and $\bm{\varepsilon}$ is a noise vector such that each entry $\varepsilon_n \sim \mathcal{N}(0,\sigma^2)$, with $n= 1,\dots,N$. Even though we assume normality, it is possible to assume other distributions for the noise vector. Moreover, it is feasible to consider a heteroscedastic structure for some components of the model, as will be presented in Section \ref{case}.

In Equation (\ref{bmodel}), each vector $\bm{b}^{(1)},\bm{b}^{(2)},\ldots,\bm{b}^{(V)}$ consists of $B_1,B_2,\ldots,B_V$ values, respectively, each of which represents the levels of a factor (e.g., 8 genotypes, 10 environments and 4 soil types would yield $B_1=8,B_2=10,B_3=4$, with $V=3$). The cumulative direct sum operator $\dplus$ then ensures sums of main effects representing all possible combinations between levels, each corresponding to one observation in the data set. The additive term represents the individual effect of each predictor, while the summation captures (via $Q$ components) the interactions between the individual effects. In the case where there is only the effect of two variables, the model in Equation (\ref{bmodel}) is reduced to the AMMI model. The summation term provides a regularisation on the complexity of the model, with larger $Q$ yielding a more complex set of interactions. 
 
The model in the form presented in Equation (\ref{bmodel}) allows for the inclusion and study of multiple categorical predictors beyond the standard $G\times E$ pair used in AMMI models, and the understanding of their effects in two parts, individually and when interacting. As in the traditional AMMI model, $Q$ is fixed and represents how many multiplicative terms are included in the model. Common extra predictors that might be added to the model include soil type, time, or growth stages, amongst many others. Being able to tractably estimate the effect of each of these on a phenotype would be extremely useful for practitioners, whilst retaining the simple interpretation of the parameters in the AMMI model. 

Importantly, our proposed model allows the inclusion of the entirety of potential interactions, from 2-way to $V$-way. Through the use of appropriate selection priors and by adjusting the value of the hyperparameter $Q$, we ensure that the model not only accounts for these interactions but also scales efficiently with the complexity introduced by them. Thus, interactions between certain factors that are not present in the data are not incorporated. This is a key feature of the model, as by allowing for the modelling of multiple levels of interactions, the model retains predictive and explanatory power. This can be useful when we want the model to be flexible and adapt to the data, or when we have prior knowledge that the effect of an interaction should be negligible.



\subsection{Prior distributions in the BAMMIT model} \label{inference}




In order to ensure the tractability of the coefficients in the model, it is necessary to establish restrictions on the interaction terms, similarly to those applied in the AMMI model. However, it is not trivial to ensure the identifiability of each parameter individually, only the entire product term \citep{guhaniyogi2017bayesian}. In the Bayesian context, these constraints are ensured from the definition at the prior level. For example, in the Bayesian AMMI model proposed by \cite{perez2012general}, the von Mises-Fisher distribution is considered for the coefficients of the multiplicative term. In the tensor field, \cite{guhaniyogi2017bayesian}  introduce multiway shrinkage priors in their tensor regression model. In our approach, we provide a new method by which the constraints are met by applying the restrictions of the interaction term through parameter transformations which we describe next. 

Formally, we frame a hierarchical model in which prior distributions of the grand mean, main additive effects and variance parameters are 
\begin{eqnarray*}
\mu &\sim& \mbox{N}(\mu_\mu, \sigma^2_\mu), \\     
\bm{b}^{(v)} &\sim& \mbox{N}(0,\sigma^2_{\bm{b}^{(v)}}),  \\
\lambda_q &\sim& \mbox{N}^+(0,\sigma^2_\lambda),\\   
\sigma^{-2} &\sim& \mbox{G}(a_0,a_1),\\
\sigma_{\bm{b}^{(v)}} &\sim& \mbox{t}^+(0,a_2),\\
\sigma_\lambda &\sim& \mbox{t}^+(0,a_3), 
\end{eqnarray*}
where $\mbox{N}$, $\mbox{N}^+$, $\mbox{G}$, and $\mbox{t}^+$, are the Normal, truncated Normal, Gamma, and truncated $t$-Student distributions, respectively. The hyperparameters of the grand mean $\mu_\mu$ and $\sigma^2_\mu$ are fixed as are all $a_k$ terms, $k = 0,1, 2, 3$. We treat the additive effects as random and so estimate $\sigma_{\bm{b}^{(v)}}$, though a `fixed effects' version could also be implemented. We express the prior knowledge on the standard deviations of the additive term parameters and the $\lambda$ parameter using a truncated $t$ distribution. Additionally, we impose that the estimated
$\bm{\lambda}$ vector values are in descending order.

For the product parameters in the interaction term, we use a transformation to ensure the constraints are met and to capture all the interaction orders. Specifically, we generate an auxiliary variable $\theta_{\beta^{(v)}_{iq}}$ from a standard $\mathrm{N}(0,1)$ distribution (the transformation is invariant to the scale of this distribution), with $i = 1, \dots, B_v$, $v = 1, \ldots, V$, $q = 1, \dots, Q$. Then, we centre by the mean $\mu_{\bm{\beta}^{(v)}_{q}}$ of the vector $\bm{\beta}^{(v)}_{q}$, that is, for each vector $\bm{\beta}^{(v)}_{q}$ we calculate its mean and then subtract it from the auxiliary variable $\theta_{\beta^{(v)}_{iq}}$ for the respective value of \textit{q}. Finally, we get $\beta^{(v)}_{iq}$ via:
\begin{eqnarray}\label{const1}
\beta^{(v)}_{iq} &=& \left(\theta_{\beta^{(v)}_{iq}} - \mu_{\bm{\beta}^{(v)}_{q}}\right)\left[\sum_i \left(\theta_{\beta^{(v)}_{iq}} - \mu_{\bm{\beta}^{(v)}_{q}}\right)^2\right]^{-1/2} .
\end{eqnarray}

Applying this procedure to the parameters of the matrix $\bm{\beta}^{(v)}$ guarantees that the identifiability constraints (2) and (3) of the model are met in the inferential process. To ensure individual variables can be removed from the interaction terms (and hence capture sub $V$-way interactions) we include extra parameters in the prior to give:
\begin{eqnarray*}
\beta^{(v)}_{iq} &=& \left[\left(\theta_{\beta^{(v)}_{iq}} - \mu_{\bm{\beta}^{(v)}_{q}}\right) + M \gamma^{(v)}_q\right]\left[\left(\sum_i \left(\theta_{\beta^{(v)}_{iq}} - \mu_{\bm{\beta}^{(v)}_{q}}\right)^2\right)^{1/2} + M \gamma^{(v)}_q \right]^{-1},
\end{eqnarray*}
\noindent
where the binary variable $\gamma^{(v)}_q$ acts as a switch in the model, allowing the model to include or exclude the effects of the corresponding variable $v$ depending on the data. The value $M$ is set as a large constant. Thus  $\gamma^{(v)}_q = 1$ implies $\beta^{(v)}_{iq} \approx 1$ and so that interaction is removed. When $\gamma^{(v)}_q = 0$ the interaction is included as normal in the BAMMIT model. To achieve this, we impose the additional prior distributions:
 \begin{eqnarray*}
\gamma^{(v)}_q  &\sim& \mbox{Bernoulli}(p^v_q),\\
p^{(v)}_q &\sim& \mbox{Beta}(1,10).
 \end{eqnarray*}
 \noindent
Our approach here is akin to a  spike-and-slab structure where we can control the inclusion or exclusion of certain variables based on the data, and consequently determine which interactions are involved. We fix the values 1 and 10 in the Beta distribution to provide an approximate 10\% chance that any given interaction is important \textit{a priori}. The 95\% CI of this Beta distribution is (0.003, 0.309).

\section{Simulation Studies} \label{simulation}

In this section we examine the performance of our proposed methodology in three different ways, using both in-sample and out-of-sample validation for all scenarios. First we assess whether the proposed priors are indeed capturing the lower and higher order interactions. Then, we examine whether the inclusion of additional variables in the model truly improves prediction and observe how the model behaves with the insertion of these new variables. Finally, we conduct a comparative analysis between our method and other existing methodologies. 

\subsection{Simulation scenarios}


The simulation scenarios were designed considering $V \in \{2,3,4\}$. In each scenario, we configure the number of levels ($B_1, ..., B_V$) to allow for differences in the interaction structures. We set up the values as follows:

 \begin{itemize}
    \item[(i)] $V = 2$ and $N = 120$, with $B_1 = 12$, $B_2 = 10$. This is the classic AMMI approach with only two variables.
    \item[(ii)] $V = 3$ and $N = 480$, with $B_1 = 12$, $B_2 = 10$, $B_3 = 4$. This is an extension of the AMMI approach with three variables. This scenario is chosen to illustrate the model's ability to capture multiple levels of interaction (i.e. 2 and/or 3-way). Specifically, the interaction term of Equation (\ref{bmodel}) is partitioned into the following equations, where $I_{ij}$ and $I_{ijk}$ are used to represent the 2 and 3-way interactions of interest, respectively:
    \begin{eqnarray}
    (a) & & I_{ij} = \sum_{q=1}^{Q} \lambda_{q}\left(\beta_{i q}^{(1)} \beta_{j q}^{(2)} \right), \label{int_2_way} \\
    (b) & & I_{ijk} =  \sum_{q=1}^{Q_1} \lambda_{1q}\left(\beta_{i q}^{(1)} \beta_{j q}^{(2)}\right) +  \sum_{q=1}^{Q_2} \lambda_{2q}\left(\beta_{i q}^{(1)} \beta_{j q}^{(2)} \beta_{k q}^{(3)}\right), \label{int_3_way}\\
     (c) & & I_{ijk} =  \sum_{q=1}^{Q} \lambda_{q}\left(\beta_{i q}^{(1)} \beta_{j q}^{(2)} \beta_{k q}^{(3)}\right) \label{int_full}.
\end{eqnarray}

    \item[(iii)] $V = 4$ and $N = 960$, with $B_1 = 12$, $B_2 = 10$, $B_3 = 4$, $B_4 = 2$. This is a further extension with four variables which allows for potentially many complex and multi-layer interactions. In this case, we are not partitioning the interaction term of Equation (\ref{bmodel}). Therefore, the data is simulated while solely taking into account the 4-way interaction.
\end{itemize}




We specifically use scenario (ii) to assess whether the proposed priors are effectively capturing both lower and higher-order interactions. We chose to examine scenario (ii) because it offers a simpler context than scenario (iii) for this evaluation, and it is not limited to a 2-way interaction, as in scenario (i). The rationale behind considering sub-cases (a) and (b) in scenario (ii) is to discern if our model can recover the lower-order interactions even when we fit a model with a higher order interaction term. Similarly, we aim to investigate the effect of the parameter $Q$ when simulating the model with varying orders of interactions. Upon validating that the model successfully captures interactions at multiple levels, we then consider  simulation scenarios (i), (ii) part (c), and (iii) to investigate the impact of incorporating additional predictors into the model.



To specify the number of terms for the interaction, we define $Q_{\text{sim}}$ as the value of $Q$ used in the simulation, which can take values from the set $\{1,2,3\}$. Correspondingly, we set $\lambda$ to values given in Table \ref{tab:lambda}. For scenario (ii) part (b) we note that there are two $Q$ values (one for each of the latent interactions), so we set the values of $\lambda$ for both $Q_{1, \text{sim}}$ and $Q_{2, \text{sim}}$ as they are presented in Table \ref{tab:lambda}. For the sake of simplicity, we assume $Q_{1, \text{sim}} = Q_{2, \text{sim}}$ in all cases.

\begin{table}[H]
    \centering
    \begin{tabular}{c|c} \hline 
         $Q_{\text{sim}}$& $\lambda$\\ \hline 
         1& \{10\}\\  
         2& \{8,10\}\\ 
         3& \{8,10,12\}\\ \hline
    \end{tabular}
    \caption{Values of $\lambda$ for each $Q_{\text{sim}}$.}
    \label{tab:lambda}
\end{table}

\noindent
Specifically, considering the configuration of this scenario, there are 480 observations and three predictors, setting 12 levels for the first predictor, 10 for the second and 4 for the third. Given this number of observations and variables, we generate three configurations of training data sets: one where the value of $Q_{1, \text{sim}} = Q_{2, \text{sim}} = 1$; another where $Q_{1, \text{sim}} = Q_{2, \text{sim}} = 2$; and finally, a data set in which $Q_{1, \text{sim}} = Q_{2, \text{sim}} = 3$; where the $\lambda$ values correspond to their respective values in Table \ref{tab:lambda}. The same understanding extends to generate the corresponding test data in all scenarios. In total, we generate 30 simulated data sets (15 for training and 15 for testing). In all scenarios, we set $\mu = 100$, and $\sigma = 1.5$. The settings and model parameterisations for our current simulation study are derived from previous simulation experiments on similar models in the literature  \citep{josse2014another, sarti2023bayesian}.


To fit the BAMMIT model, we run a Markov chain Monte Carlo (MCMC) algorithm through the probabilistic programming language Just Another Gibbs Sampler \citep[JAGS;][]{plummer2003jags} and the R package \texttt{R2jags} \citep{su2021r2jags}. We fit five BAMMIT models to each simulated data set, by varying $Q\in\{1,2,3,4,6\}$. We use $\mu_\mu = 100$, $\sigma^2_\mu = 10$, $a_0 = a_1 = 0.1$, $a_2 = a_3 = 1$, and $M = 10,000$. We use three chains, 4,000 iterations per chain, discarding the first 2,000 as burn-in, and a thinning rate of two. Regarding computational time, a data set with three predictors ($V = 3$), $N = 100$ and $Q = 1$ takes on average three minutes to run, whilst to run a data set with the same configuration, but $Q = 3$ takes 17 minutes. We discuss computational issues further in Section \ref{discussion}.  All experiments were implemented in R\footnote{The code used is available at \url{https://github.com/Alessandra23/bammit}} on a MacBook Pro with a 1.4 GHz Quad-Core Intel Core i5 and 8 GB of RAM.

To assess the performance of the model when $ V>2$ (and so standard AMMI cannot be applied), we compare BAMMIT with two models extensively employed for prediction purposes, namely Random Forests (RF) and eXtreme Gradient Boosting (XGB). A Bayesian factorial mixed (BFM) model \citep{gelman2007data, fong2010bayesian} is also fitted for comparison. Ultimately, we compare our approach with the traditional AMMI and the more recent AMBARTI model \citep{sarti2023bayesian}, though these are unavoidably restricted to using only the first two variables. For the RF model, we use the package \texttt{randomForest} \citep{liaw2002classification} selecting the default settings, \texttt{mtry}$= 2$ and 500 trees. For the XGB model, we use the package \texttt{xgboost} \citep{chen2019package} setting 50 iterations. For the AMBARTI model we use the package \texttt{AMBARTI} \footnote{The code is available at \url{https://github.com/ebprado/AMBARTI}.} setting 50 trees, 500 as burn-in and 1000 iterations as post burn-in. For the BFM model, the main effects are included, along with the insertion of interactions of all orders, and we follow the same priors as applied to the BAMMIT model. All the models were fitted to the training data. We checked the accuracy, using the test data, by comparing the posterior mean estimates with the true parameter values used in the simulations. We use the root mean squared error (RMSE) to measure predictive power (how close $\hat{y}$ is to the true $y$) and $R^2$ to assess the proportion of explained variability.

\subsection{Simulation results}

\begin{figure}[!ht]
\label{int_2way_bammit}
			\centering
			\includegraphics[width=\columnwidth]{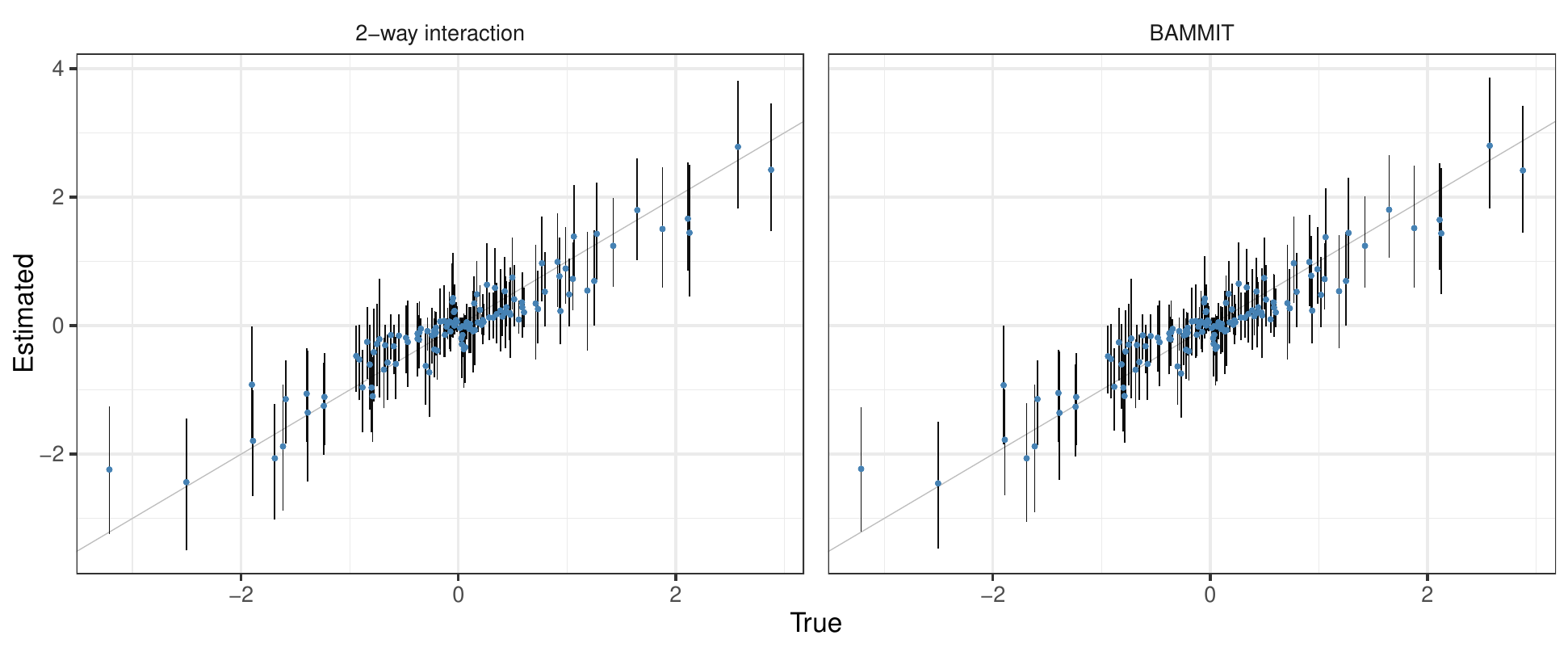}
			\caption{In-sample scatterplots of true versus estimated interaction term for simulation scenario (ii) part (a), setting $Q_{\text{sim}} = 1$ and $\lambda = 10$. The left panel shows only the estimated interaction effects whilst the right panel shows the full estimated fitted values. The models were fitted with $Q = 1$.  The blue points represent the posterior median and the grey bars represent the $95\%$ credible intervals.} 
			\label{fig:int_2way_bammit}
		\end{figure}

Figure \ref{fig:int_2way_bammit} presents scatterplots of true versus estimated values in the case where we simulated the interaction structure as given in scenario (ii) part (a), setting $Q_{\text{sim}} = 1$. The first plot corresponds to the fit of the model that follows the same structure as the simulated data, which considers all main effects along with the true simulated 2-way interaction. This model includes solely the step of imposing identifiability constraints, as outlined in Equation (\ref{const1}). The second corresponds to the BAMMIT model fit. We observe that both fits exhibit similar results, as shown by the alignment of true versus estimated interaction terms in the scatterplots, indicating that the BAMMIT structure captures the interaction between the two variables.

\begin{figure}[!ht]
			\centering
		\includegraphics[width=\columnwidth]{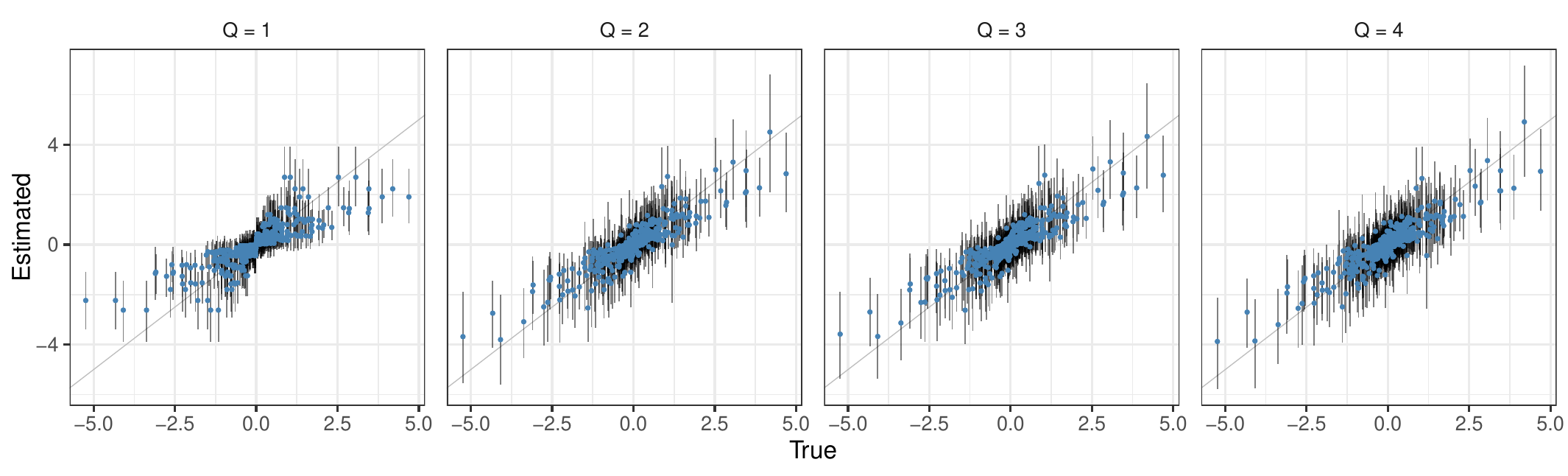}
			\caption{Scatterplots of true versus estimated interaction term for simulation scenario (ii) and Equation (\ref{int_3_way}), setting $Q_{1, \text{sim}} = Q_{2, \text{sim}} = 1$ and $\lambda = 10$. The model was fitted with $Q = \{1,2,3,4\}$. The blue points represent the posterior median and the grey bars represent the $95\%$ credible intervals.} 
			\label{fig:int_bammit_Q}
		\end{figure}

Figure \ref{fig:int_bammit_Q} illustrates the performance of the BAMMIT model under scenario (ii) part (b), where the interaction structure was simulated with $Q_{1, \text{sim}} = Q_{2, \text{sim}} = 1$, while the model was fitted using values of  $Q = \{1,2,3,4\}$. The graphs clearly highlight the importance of the value assigned to the hyperparameter $Q$ on model accuracy. Specifically, when only one term is present in the interaction ($Q = 1$), the estimation is not accurate. This is because with $Q = 1$, the model cannot capture the complexity of the two different interactions present in the simulated data. However, as $Q$ is increased the BAMMIT model successfully captures the simulated interaction structure. The lesson here is that $Q$ needs to be sufficiently large to capture all possible latent interactions in the data. 

To assess the behaviour of the BAMMIT model when data is simulated with a higher value of $Q$, we present in Table \ref{rmse_int} the RMSE of the interaction for the BAMMIT model fit, varying the number of terms in $Q = \{1,2,4,6\}$. The interaction structure was simulated again as given in scenario (ii) part (b), with the assumption that $Q_{1,\text{sim}} = Q_{2,\text{sim}} = \{1,2,3\}$. Once more, we observe that the incorporation of more terms into the interaction enhances the model's accuracy, leading to a decrease in the RMSE of the interaction as $Q$ increases.

\begin{table}[!ht]
\centering
\begin{tabular}{ccccl}
\hline
\multicolumn{1}{l}{} & \multicolumn{4}{c}{Fitted $Q$}  \\ \cline{2-5} 
$Q_{1,\text{sim}}$ ,  $Q_{2,\text{sim}}$     & 1 & 2 & 4 & 6 \\ \hline
 1                & 2.75  & 0.76  & 0.59  & 0.58     \\ \hline
 2                & 3.38  & 2.51  & 1.21  & 0.98  \\ \hline
3                & 5.65  & 2.47  & 0.88  & 0.84  \\ \hline
\end{tabular}
\caption{In-sample RMSE for the interaction term of the BAMMIT model, considering that the interaction was simulated considering case (ii) part (b) and $Q_{1,\text{sim}} = Q_{2,\text{sim}} = \{1,2,3\}$. The BAMMIT model was fitted with $Q = \{1,2,4,6\}$. We would expect the model to perform satisfactorily once the fitted $Q$ value is greater than or equal to $Q_{1,\text{sim}} + Q_{2,\text{sim}}$.}
\label{rmse_int}
\end{table}

\begin{figure}[!ht]
			\centering
			\includegraphics[width=\columnwidth]{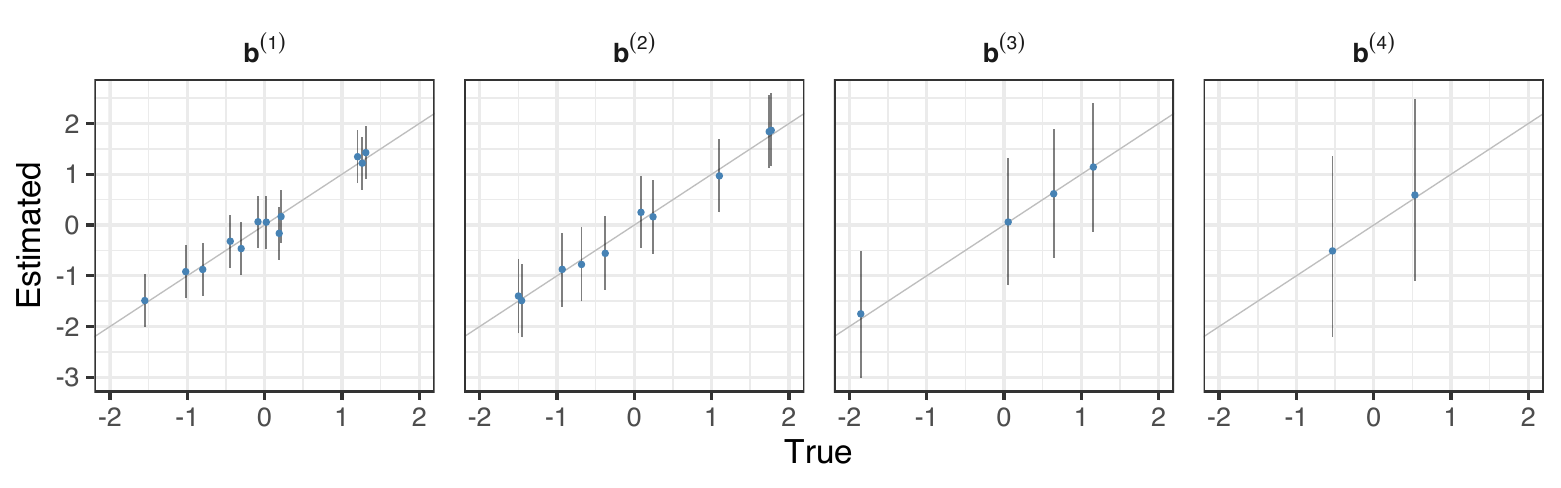}
			\caption{Scatterplots of true versus estimated additive terms for simulation scenario (iii), setting $Q_{\text{sim}} = 2$, $\lambda = \{8,10\}$. The blue points represent the posterior median and the grey bars represent the $95\%$ credible intervals.} 
			\label{fig:mainEffV4}
		\end{figure}

Given that the model is successfully capturing both the lower and higher  interactions, we now focus on presenting the results where the data simulation was conducted directly from Equation (\ref{bmodel}), and how the inclusion of new terms contributes to the model's prediction. Initially, the scatterplot in Figure \ref{fig:mainEffV4} shows the comparison of the additive term, taking $V = 4$, $N = 960$, and $Q_{\text{sim}} = 2$,  with true value of $\lambda = \{8,10\}$. The models was fitted with $Q = 2$. Each point is an estimated value of the parameters and the error bars are the $95\%$ credible intervals. By visual inspection, the estimates of the effects of the four main predictors are close to the true values, with narrower intervals for predictors with a greater number of levels. 

\begin{figure}[!ht]
			\centering			\includegraphics[width=\columnwidth]{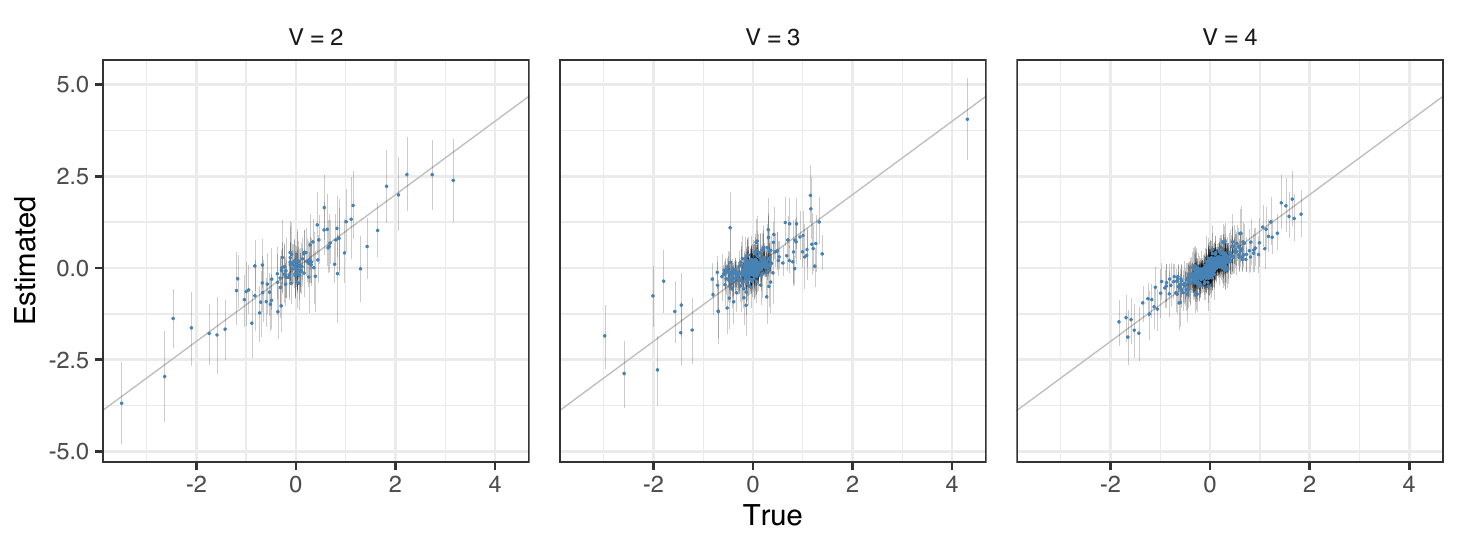}.			\caption{Scatterplots of true versus estimated interaction terms for simulations scenarios (i), (ii) and (iii) setting $Q_{\text{sim}} = 1$ and $\lambda = \{10\}$. The interactions generated were 2-way ($V = 2$), 3-way ( $V = 3$) and 4-way ($V = 4$). The blue points represent the posterior median and the grey bars represent the $95\%$ credible intervals.}
			\label{fig:blinVQ3}
		\end{figure}

In Figure \ref{fig:blinVQ3}, we compare the estimates against the true values in the case where the number of predictors varies. Each point represents an interaction term estimate in a total of 120 ($V=2$), 480 ($V=3$) and 960 ($V=4$) points, and the bars, again, represent the $95\%$ credible intervals. We observe that when $V = 4$, the dispersion is smaller and the interaction estimates are more concentrated around zero. This can be explained because as more predictors are added to the additive term of the model, the greater the approximation of the response by the predictors and the smaller the amount approximated by the interaction term, despite inserting more variables in both terms of the model. Also, note that the interaction is comprised of all the new variables together, and that this interaction may not be that strong. For example, suppose we are looking at the \textit{genotype} $\times$ \textit{environment} $\times$ \textit{soil type} $\times$ \textit{growth stage} interaction. In this case, the interaction of the four factors together is not as strong as if we were looking only at subsets of these interactions, such as \textit{genotype} $\times$ \textit{environment} $\times$ \textit{growth stage}.

In terms of predictions, Table \ref{rmsesim} shows the prediction RMSE and the $R^2$ considering the cases where we have three and four predictors in the models (simulation scenarios (ii) and (iii)). 
To fit BAMMIT and AMMI models we used $Q=2$. As stated above, the AMBARTI and AMMI models were fitted disregarding the effects of the other variables. Specifically, in scenario (iii), for example, there were three predictors, but the two aforementioned models disregarded the effect of the third predictor. The BAMMIT model clearly performed better than the other two models. In addition to the prediction advantage, our model stands out from RF and XGB as it can provide a simpler structure for the interaction between the variables, while at the same time providing estimates based on full posterior distributions rather than point estimates. Our approach was better than the BFM, that includes all the possible interactions between the variables. Our BAMMIT model was able to satisfactorily explain the variability of the response variable, since the $R^2$ obtained in all scenarios was above 75\%. In a real world scenario where the data were not simulated from the BAMMIT model we might expect that the machine learning approaches would be more competitive in terms of their performance. However, they would still not allow for clear interpretation of the interaction effects.


\begin{table}[!ht]
\footnotesize
\makebox[\textwidth][c]{%
\begin{tabular}{ccccccc|cccccc}
\cline{2-13}
\multicolumn{1}{l}{} & \multicolumn{6}{c|}{V = 3}         & \multicolumn{6}{c}{V = 4}          \\ \cline{2-13} 
                     & BAMMIT & AMBARTI & AMMI & RF & XGB & BFM & BAMMIT & AMBARTI & AMMI & RF & XGB & BFM \\ \hline
RMSE                 &     \textbf{1.62}    &     2.54    &  2.52    &   1.68    &   2.06  & 1.87  &    \textbf{1.64}     &     2.74    &   2.71   &    1.74    &    2.05  & 1.72 \\
$R^2$                &   \textbf{0.78}  &    0.02     &   0.02   &    0.58   &  0.69  & 0.72   &    \textbf{0.81}   &    0.01     &   0.01   &    0.68    &   0.78   & 0.79 \\ \hline
\end{tabular}}
\caption{RMSE and $R^2$ for $\hat{y}$ on \textit{out-of-sample} data for scenarios (ii) and (iii), setting $Q_{\text{sim}} = Q = 2$.}
\label{rmsesim}
\end{table}




\section{Case Study} \label{case}

In this section, we investigate the performance of the model on a real data set. The data was collected over ten years (2010 -- 2019) and concerns the production of a common species of wheat (\textit{Triticum aestivum L.}) in Ireland, with the response being the yield of wheat measured in tonnes per hectare (\textit{t/ha}). The data comes from the Horizon2020 EU InnoVar project\footnote{\url{www.h2020innovar.eu}} and was supplied by the Irish Department of Agriculture, Food and the Marine. The experiments were conducted using a randomised complete block design with four replicates. The data set contains 85 genotypes and 17 environments, all anonymised and named as $g_1,\dots, g_{85}$ and $e_1,\dots, e_{17}$, respectively. Here, environment refers only to the location, instead of the common interpretation of environment as a location-year combination. Owing to not all genotypes being observed in each location in all seasons, the total number of observations \textit{genotype $\times$ environment $\times$ year $\times$ block} is 6,368, rather than 57,800. It exemplifies one of the advantages of the BAMMIT model, that is able to impute the missing combinations as part of the model fit.

A subsample of this data was previously explored by \cite{sarti2023bayesian}, considering only two factors: genotype and environment. However, in our work, we include the additional variables \textit{year} and \textit{block}, present in the Irish data set, as a third and fourth effect in the BAMMIT model. We expect to detect if there is variability between years across environments and genotypes, and to determine such an interaction. Previously \cite{hara2021selection} showed that the ability to predict the yield in a certain year can be useful for making decisions such as cultivation planning and storage. Specifically, the model structure for the data is
\begin{eqnarray}\label{real_data_bammit}
y_{ijtr} &=& \mu + b^{(1)}_i + b^{(2)}_j + b^{(3)}_t  + b^{(4)}_{jr} + \sum\limits_{q=1}^Q \lambda_q\beta^{(1)}_{iq}\beta^{(2)}_{jq}\beta^{(3)}_{tq}+\epsilon_{ijtr},
\end{eqnarray}
\noindent
where the indexes $i, j, t$ and $r$ are associated to genotypic, environmental, time and block effects, respectively. We nest the block effect within environment and allow an environment-specific variance for each, thus removing any experimental variation associated with that specific site not related to the response. To avoid numerical under- or overflow we standardise the response before fitting, but convert all predicted values back to the original scale for ease of interpretability. 


To fit the model, we partition the data into training and testing sets by selecting three out of the four available blocks for training. The selection of this number of blocks is to ensure a more comprehensive representation of the inherent variability across blocks. The remaining block is used for validation, resulting in 4,776 observations for training and 1,592 for testing. Therefore, we have $V = 4$, $B_1 = 85$ genotypes, $B_2 = 17$ environments, $B_3 = 10$ years and $B_4 = 3$ blocks. Assuming that there is little prior information for hyperparameter specification, we follow a non-informative approach, with $a_0 = a_1 = 0.1$. 
To input the number of $Q$ terms, we run the model with $Q = \{1, 2, 3,4\}$. The model is fitted with three Markov chains, 4,000 iterations per chain, discarding 2,000 and a thinning rate of two. 

We compare our BAMMIT approach prediction performance with traditional AMMI, AMBARTI and a Bayesian factorial model. For the AMBARTI model, we consider 50 trees, 1000 iterations as burn-in and 1000 iterations as post burn-in. The fitted Bayesian factorial model we use follows the same priors assumed for the BAMMIT model. We define the model structure as:
\begin{eqnarray*}
    y_{ijtr} = \mu +  b^{(1)}_i + b^{(2)}_j + b^{(3)}_t + b^{(4)}_{jr}+ b^{(1,2)}_{ij} + b^{(1,3)}_{it} + b^{(2,3)}_{jt} + b^{(1,2,3)}_{ijt} + \epsilon_{ijtr}.
\end{eqnarray*}

We note that this model no longer parameterises the interactions in a latent tensor space (as our models do) and so uses a far greater number of parameters. Since we only evaluate our model on out of sample data the comparison between these approaches remains valid. 

In addition to evaluating the model's performance in terms of prediction, we are interested in answering some specific questions. Initially, when fixing the year and block effect, we would like to know which genotype has the best performance, in which environment, and also which environment provides the highest yield. When considering all the variables, we investigate which year has the best performance. Subsequently we explore the interactions between these variables to find potential year/environment/genotype combinations which provide optimal or sub-optimal yield performance.

\subsection{Results}

To assess the performance of the predictions $\hat{y}$, we display the RMSE and the $R^2$ in Table \ref{tab:rdrmse} for each model. As the AMBARTI and classic AMMI models can only handle the genotype and environment variables, when fitting these models we take into account all rows in the data set and ignore the year and block variables, so that rows corresponding to the same genotype and the same environment are treated as different. We considered $Q = 4$ for the BAMMIT and AMMI models. In terms of convergence, all the models have had their convergence checked, with $\hat{R} \approx 1$ \citep{gelman1992inference}. In particular, all parameters of the BAMMIT model converge, except for the individual values of the parameters $\bm{\beta}_q^{(v)}$, as expected.\\


\begin{table}[!ht]
\centering
\begin{tabular}{ccccc}
\hline
         & BAMMIT  & \makecell{AMBARTI} & \makecell{AMMI}  & BFM \\ \hline
RMSE     &     \textbf{0.59}       & 1.68      &  1.60 & 0.72 \\ 
$R^2$     &     \textbf{0.92}      & 0.36    &  0.38  & 0.85 \\ \hline
\end{tabular}
 \caption{Metrics for \textit{out-of-sample} $y$ for the BAMMIT, AMBARTI, AMMI and Bayesian factorial mixed (BFM) models. The AMBARTI and AMMI models ignored the effects of the year and the block variables. Best performance is shown in bold.}
\label{tab:rdrmse}
\end{table}

The results presented in Table \ref{tab:rdrmse} show that the BAMMIT model was superior to the other models, in both RMSE and $R^2$. Specially, our proposed model outperforms the BFM model, which is the main competitor to BAMMIT, since it directly includes all possible interactions. For the AMBARTI and the classical AMMI model, the inferior performance was expected, since the adjusted models do not take into account the effects of the other variables. Also, for the AMBARTI, these results can be explained by the high numbers of genotypes and environments. As mentioned above, considering all the 10 years, there are 85 genotypes and 17 environments, and in this case the AMBARTI model is not efficient in the generation of the $2^{B_1 - 1} - 1$ and $2^{B_2 - 1} - 1$ two-partition combinations for genotypes and environments. Thus, due to the high numbers of possible combinations, the interaction component of the AMBARTI model is not able to estimate the interactions between genotypes and environments efficiently and the model performs poorly. 

\begin{figure}[!ht]
 			\centering	\includegraphics[width=\columnwidth]{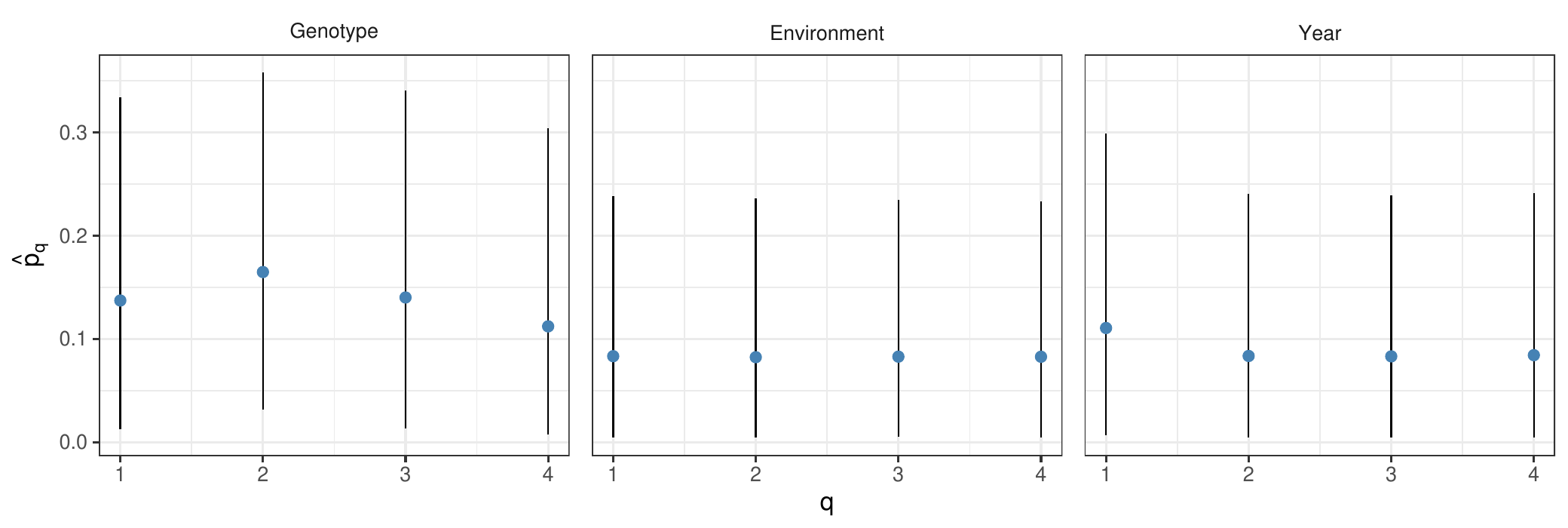}
			\caption{Estimated values of $\hat{p}^{(v)}_q$ and the 95\% credible intervals. Values closer to zero indicate an increasing probability that the variable was included in the interaction term. The general low values indicate a high degree of interaction, with environment being particularly important. We note that the uncertainty ranges in these values are far smaller than that of the $Be(1, 10)$ prior.}
			\label{p_q}
	\end{figure}

In Figure \ref{p_q}, we present the plot of the probabilities $\hat{p}^{(v)}_q$, where we observe which variables are being included in the model as well as the lower-order interactions. As observed, the model's results suggest that the effects of \textit{genotype}, \textit{environment} and \textit{year} are unlikely to be constant on wheat yield prediction. This indicates that wheat yield is influenced by the interaction of these variables at different levels, with \textit{environment} and \textit{year}  exhibiting the most interactions with the others.

\subsection{Posterior visualisation}

A common way to visualise the genotype and environment interactions in an AMMI model is through biplots \citep{gabriel1971biplot}. However, \cite{sarti2023bayesian} used a heatmap to visualise the predictions and interactions. In this visualisation approach, it is possible to identify in a more immediate way the best interactions between genotypes and environments. A shortcoming of their approach concerns the quantification of uncertainty, which cannot be observed directly on the graph. To address this particular issue, in this work we show the prediction for interactions through a heatmap as in \cite{sarti2023bayesian} and the uncertainty is showed as Value-suppressing uncertainty palettes (VSUP) as presented by \cite{inglis2022visualizations}. 

First introduced by \cite{vsup}, value-suppressing uncertainty palettes are bivariate colour palettes that represent a measure or value and its uncertainty. The outputs for each combination of value and uncertainty in traditional bivariate palettes are often shown as a 2D square \citep[for example, see][]{robertson1986generation}, However, VSUP plots combine cells in the palette using a funnel structure to suppress the measure or value at higher levels of uncertainty. In VSUP, when the uncertainty is low, more bins are allocated to the colour space. When increasing in uncertainty, the values are suppressed into fewer bins that blend together their colour value. By doing this, the values will become more distinct as the level of uncertainty reduces, with the intention of making it easier to detect the difference between low and high uncertainty.

\begin{figure}[!h]
 			\centering	\includegraphics[width=\columnwidth]{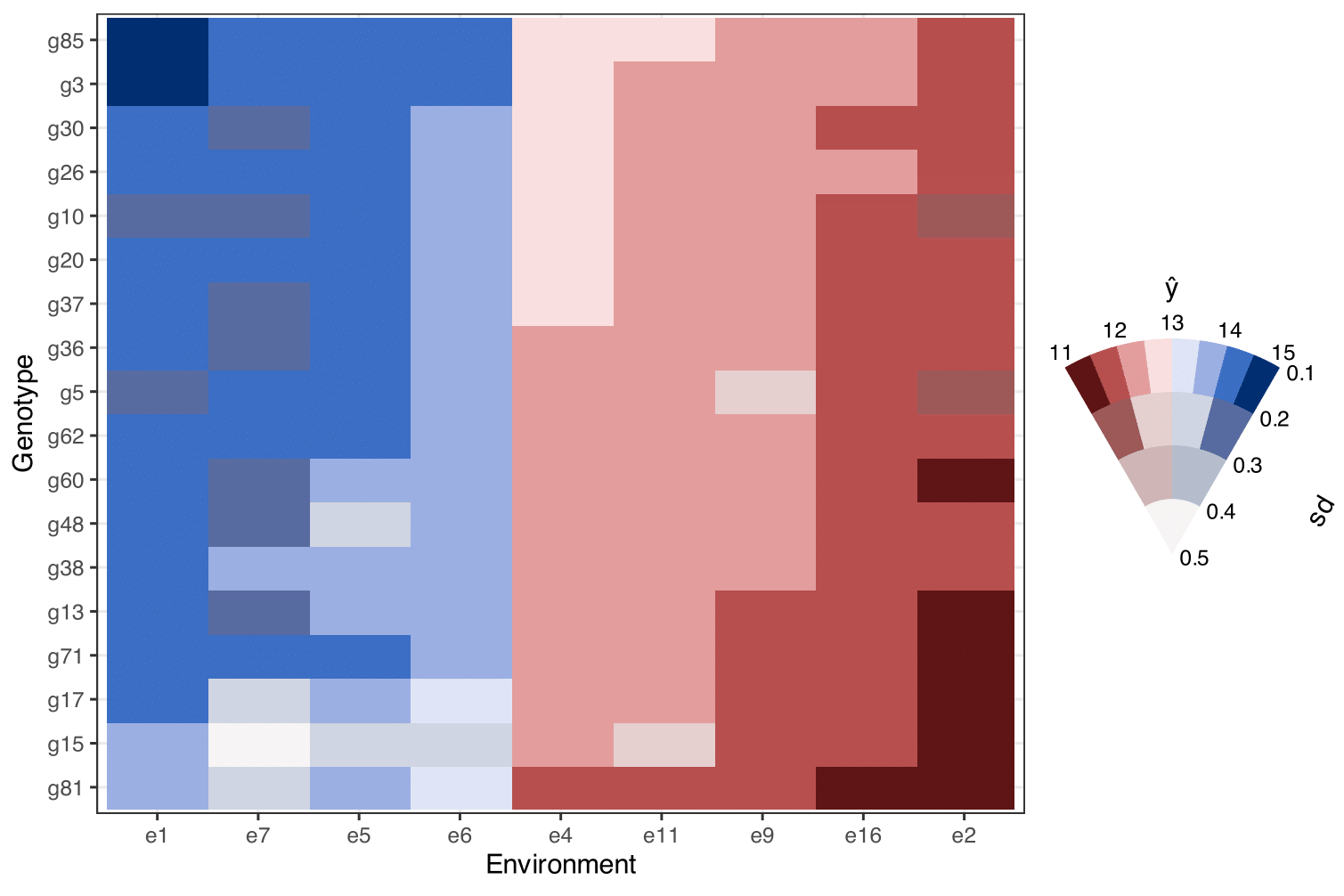}
			\caption{Predicted yields from the BAMMIT model for the wheat production data set in 2015. Production this year was high, between 11 and 15 tonnes per hectare, with positive emphasis on the environment $e_1$ and on the combinations of genotype $g_{3} \times e_{1}$ and $g_{85} \times e_{1}$.}
			\label{heatmap}
	\end{figure}

In Figure \ref{heatmap}, we display an ordered heatmap and VSUP legend for the Irish data selecting the year 2015 in block four from the test data set in the original scale. This year corresponds to the best average wheat production observed in the data. The remaining years are shown in the Appendix. For these plots we use the standard deviation of the predictions as our measure of uncertainty and the value shown is the median prediction for each variable pair. The plots are ordered so that generally high predicted yield values are pushed to the top left of the plot and descend to the bottom right. The environments $e_2$, $e_4$, $e_9$, $e_{11}$ and $e_{16}$ were the worst environments observed, having a small median value compared to the others. On the other hand, environment $e_1$ was the best and most stable, presenting a median yield value higher than the others and a lower uncertainty. The environment $e_{6}$ had a middling production across some genotypes, but with a higher uncertainty when compared to the others. Applying the same interpretation to the genotypes, we observed that the genotypes $g_{3}$ and $g_{85}$ had the best performance. The genotype $g_{10}$ presented a good production of predicted wheat on environment $e_{1}$ and $e_{7}$, however with a higher standard deviation than the other genotypes, which also produced around 15 \textit{t/ha} in this environment. As an example of the main \textit{genotype $\times$ environment} combinations, the worst observed were $g_{81} \times e_{16}$, $g_5 \times e_{2}$ and $g_{10} \times e_{2}$, while the best were $g_{85} \times e_{1}$, $g_3 \times e_{1}$, $g_5 \times e_{7}$ and $g_3 \times e_{7}$. The results are comparable to those of \cite{sarti2023bayesian}. 

The proposed VSUP plots offer a clear advantage over traditional biplots by enabling quicker pattern identification and straightforward interpretation of complex interactions. They are more scalable, handling numerous genotypes and environments without losing readability. This is a notable improvement over biplots, which become cluttered as variables increase and also require a level of expertise to interpret them. Additionally, VSUP plots include the uncertainty, which is missing from a biplot, and allows for quick identification of variable pairs with high or low confidence.

\begin{figure}[!h]
 			\centering	\includegraphics[width=\columnwidth]{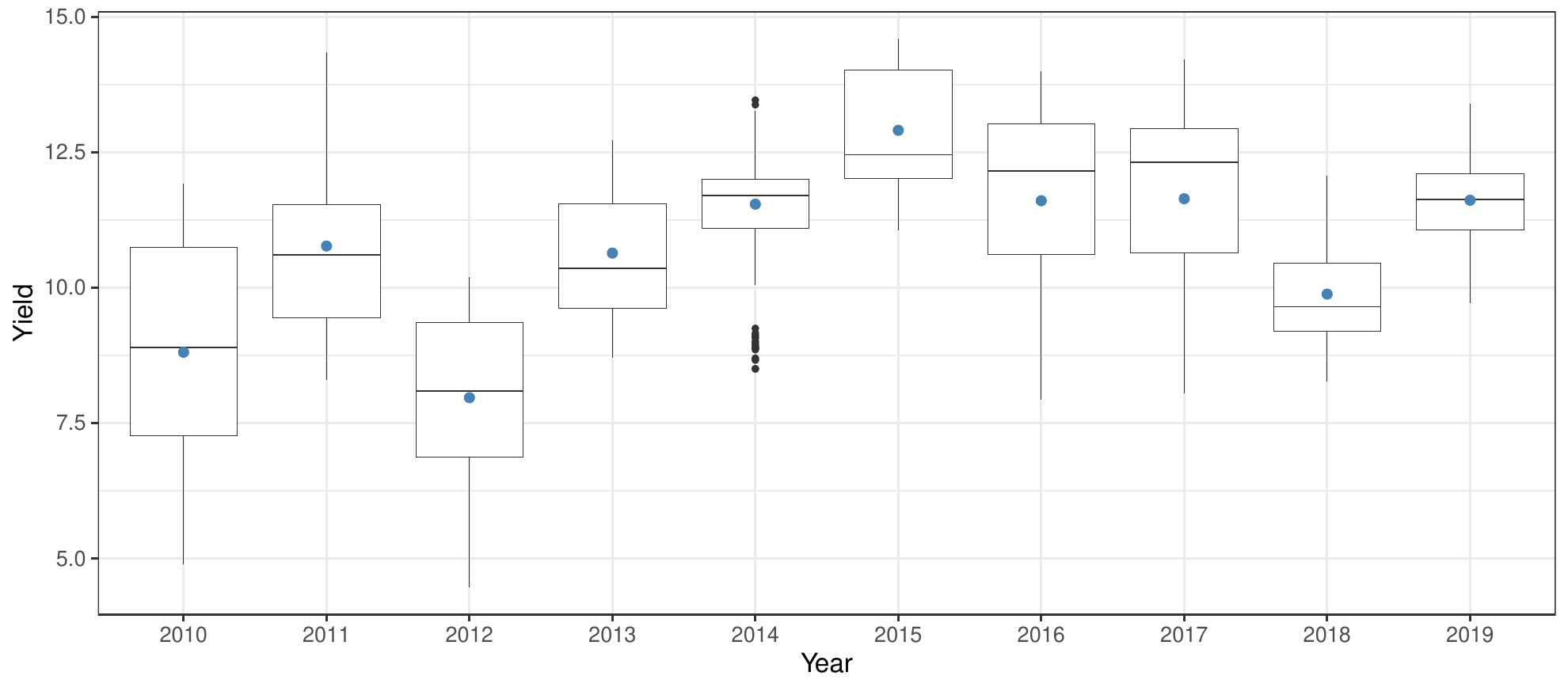}
			\caption{Box plot of the predicted wheat yield by year. The blue points represent the true mean observed in the test data.}
			\label{year_mean}
	\end{figure}

Finally, in Figure \ref{year_mean} we present the box plot of $\hat{y}$ by year, alongside the actual means observed in the test data. The true means fall within the box plots, what shows the good BAMMIT model's performance in capturing the central tendency of the wheat yield in each year. The predictions suggest that the best production occurred in the years 2015 to 2017, with 2015 having a smaller variation than the others. The analysis of the variable year in this context is particularly important, as it serves as a proxy for multiple factors that change over time, such as climatic conditions and agricultural practices, which are not explicitly captured by the static variable environment (that only represents location).


\newpage

\section{Discussion}\label{discussion}

We proposed a generalisation of the AMMI model which extends the tensor regression approach of \cite{guhaniyogi2017bayesian} and \citet{papadogeorgou2021soft} to allow for multiple interacting categorical variables. The main idea is to allow for more realistic understanding of phenotypic effects beyond the usually considered pair of genotype and environment. We envisage that in the future such models may be used to further indicate interactions between season, soil, weather conditions, growth stage and other potential predictors. The priors we use on the hierarchical model were built not only to meet the inherent restrictions and ensure identifiability but also to capture both lower and higher-order interactions between the variables.

The simulation results suggest that the model performs well in a variety of prediction tasks whilst retaining simple interpretative output. The model was able to capture the lower and higher-order interactions between the variables regardless of the true number of terms initially specified for the multiplicative term in the simulation. The results indicated that, provided the value of $Q$ was sufficiently large, the model's accuracy in capturing interactions at all levels was satisfactory. Secondly, the model demonstrated robustness when multiple interactions were included with differing levels of complexity, and the model retained the ability to estimate values of the parameters that matched the ground truth. Lastly, our model outperformed other well-known models, such as random forests and factorial mixed models in terms of predictive accuracy. As expected, in all scenarios, when the established number of terms in the interaction increases, the Bayesian model had a better fit, regardless of the true number used in the simulation. All our models were checked to confirm that the algorithm had converged and that any imposed prior constrains were met for all our simulated data sets.

When applied to real data, our approach was superior to other methods when evaluated on predictive performance and RMSE. The model enabled the identification of variables involved in interactions and determine which genotypes, environments and years had the highest wheat production. In this study, understanding the behaviour of the variable year was particularly important because the environment is limited to location and is not capable of capturing annual changes in climatic conditions or agricultural strategies, for example. Furthermore, we could visualise components of the model, such as the interaction effect, and so it was possible to determine the optimal interactions. The purpose of showing the results in the visualisations presented in this paper is to aid a researcher’s ability to interpret the results and improve recommendations. 

In relation to the computation time, the cost for the method was high, particularly as the number of predictors $V$ and components $Q$ increased. For example, a data set with three predictors ($V = 3$) and $Q = 1$ took on average one minute to run, whilst for a data set with four predictors ($V = 4$) and $Q = 3$ took around 30 minutes. 
This drawback was compounded when the data set is large
(around 5,000 total observations or more) with the model taking several hours to form a valid posterior distribution. To avoid this computational cost, potential optimisation strategies can be employed, such as parallel processing, variational inference \citep{blei2017variational, dos2022variational} or those as discussed in \cite{papadogeorgou2021soft} and \citet{zhang2020islet}.

For future work, several extensions can be made to these models. By modifying the prior distributions, new structures can be added to certain predictors, thereby allowing any temporal and spatial components to have their inherent characteristics inserted into the model. Another important extension is the insertion of continuous variables, or latent representations of them, since the current structure does not allow for this type of variable. Also, the choice of the number of  components $Q$ is still arbitrary, taking into account only the typical values already mentioned in the literature. Nonetheless, a more sophisticated approach to choosing the rank $Q$ can be applied, as shown by \cite{guhaniyogi2017bayesian}.

\section*{Acknowledgments}
Antônia A. L. dos Santos, Andrew Parnell, and Danilo Sarti received funding from the European Union’s Horizon 2020 research and innovation programme under grant agreement No 818144. Andrew Parnell’s work was supported by: a Science Foundation Ireland Career Development Award (17/CDA/4695); an investigator award (16/IA/4520); a Marine Research Programme funded by the Irish Government, co-financed by the European Regional Development Fund (Grant-Aid Agreement No. PBA/CC/18/01); SFI Centre for Research Training in Foundations of Data Science 18/CRT/6049, and SFI Research Centre awards I-Form 16/RC/3872 and Insight 12/RC/2289\_P2. For the purpose of Open Access, the author has applied a CC BY public copyright licence to any Author Accepted Manuscript version arising from this submission.

\noindent {\bf{Conflict of Interest}}
\noindent {\it{The authors have declared no conflict of interest.}}

\section*{Appendix}
\label{appendixA}


We complement the results presented in the Section \ref{case} by presenting in Figure \ref{fig:vsup_all} the graphs for the predicted yields for the BAMMIT model applied to the Irish time series data set. Analysing only the legend of the figures and looking at the value scale it is possible to see that the forecast of wheat yield for the year 2015 (presented in Figure \ref{heatmap}) was higher than that for all the  other years. In order to clearly observe the behaviour of genotype and environment predictors over the years, we scale the estimated value and the uncertainty legend to be equal across plots. Thus, years with high production and low uncertainty have an intense colour while years with contrary behaviour have a washed-out colour.

\begin{figure}[H]
\centering
\subfloat[2010]{\label{y2010}
\centering
\includegraphics[width=0.48\linewidth]{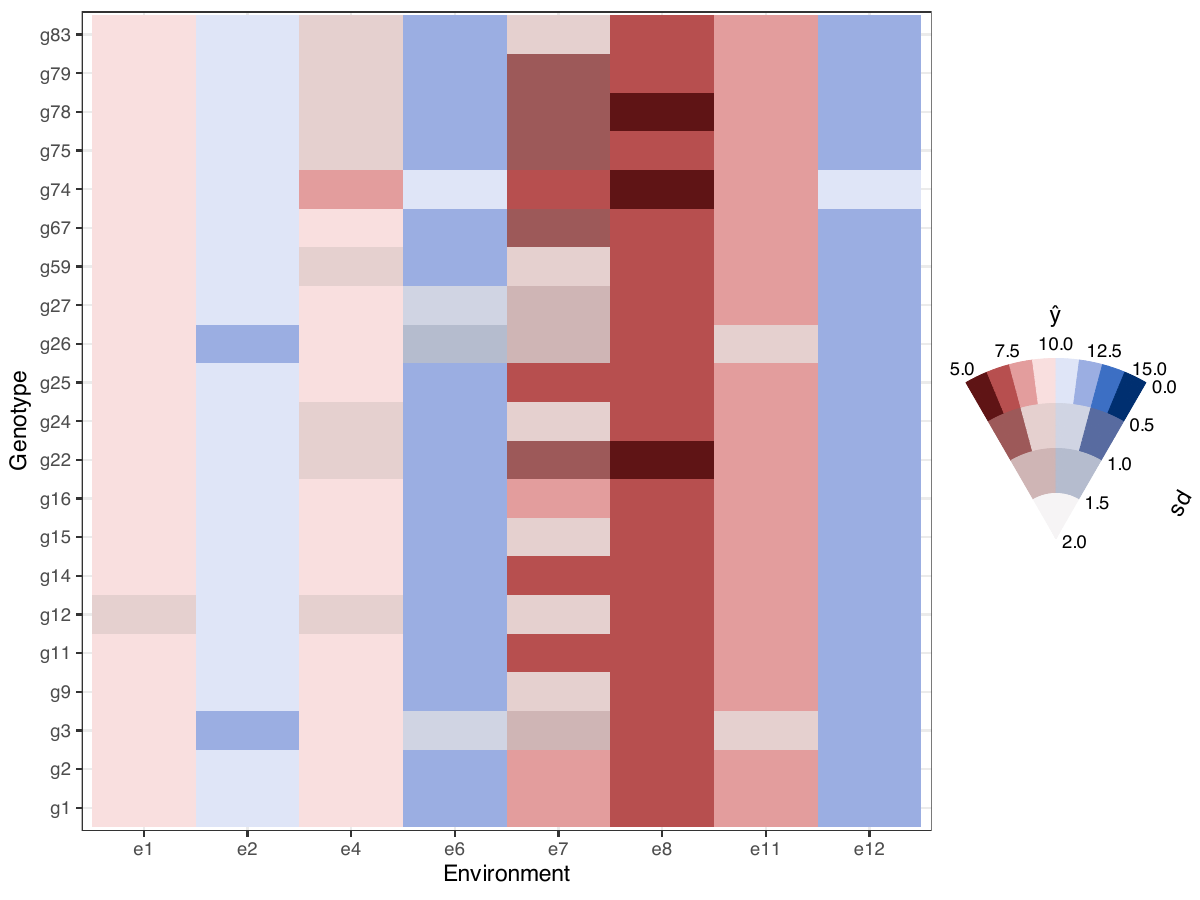}
}
\hfill
\subfloat[2011]{\label{y2011}
\centering
\includegraphics[width=0.48\columnwidth]{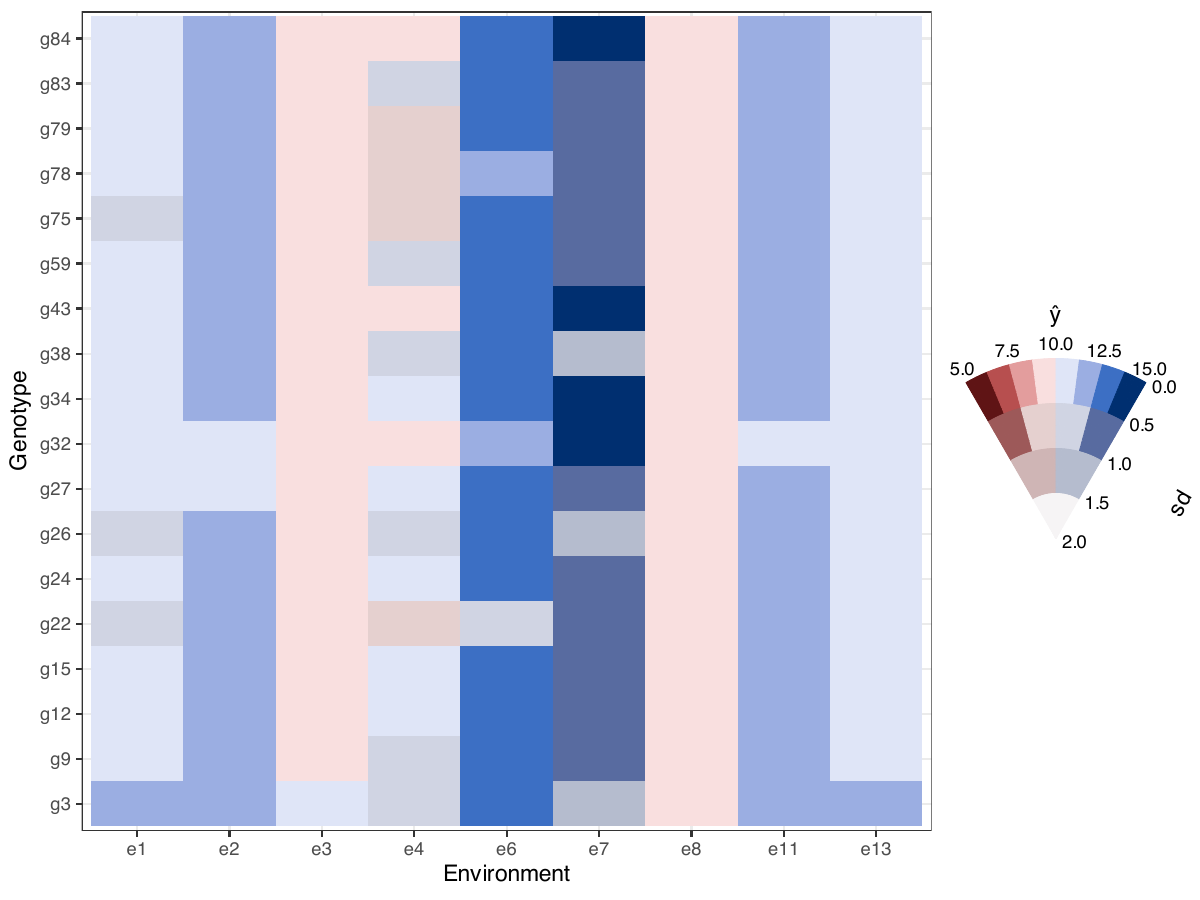}
}\\
\subfloat[2012]{\label{y2012}
\centering
\includegraphics[width=0.48\columnwidth]{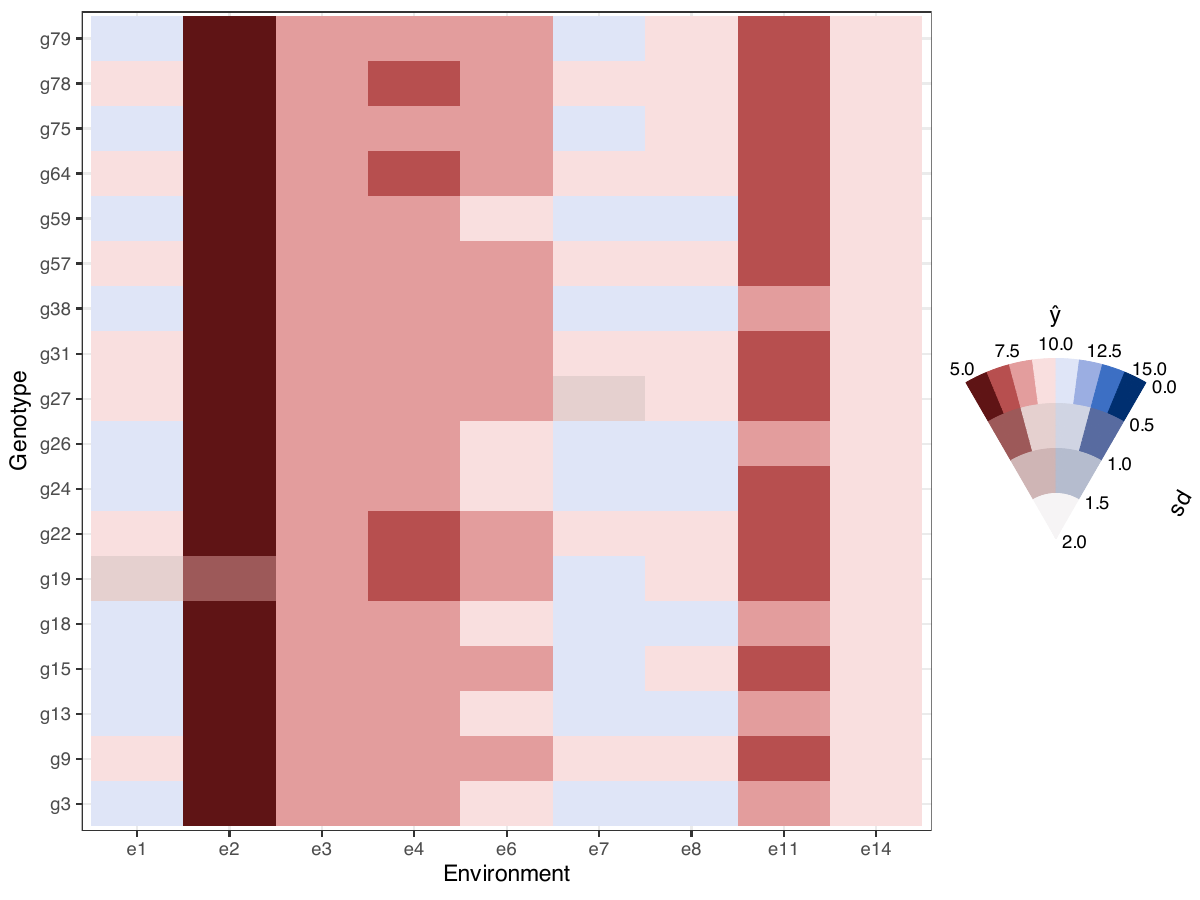}
}
\hfill
\subfloat[2013]{\label{y2013}
\centering
\includegraphics[width=0.48\columnwidth]{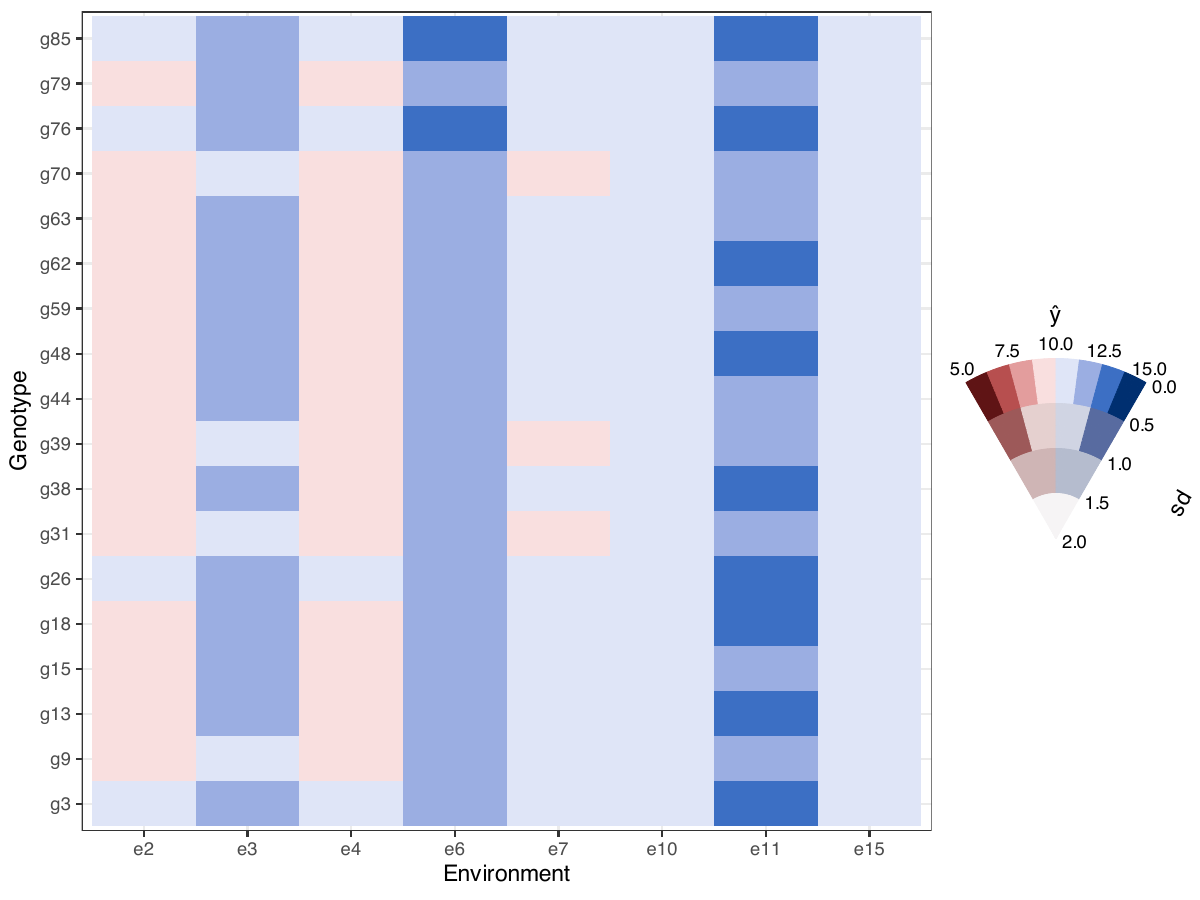}
}\\
\centering
\subfloat[2014]{\label{y2014}
\centering
\includegraphics[width=0.48\columnwidth]{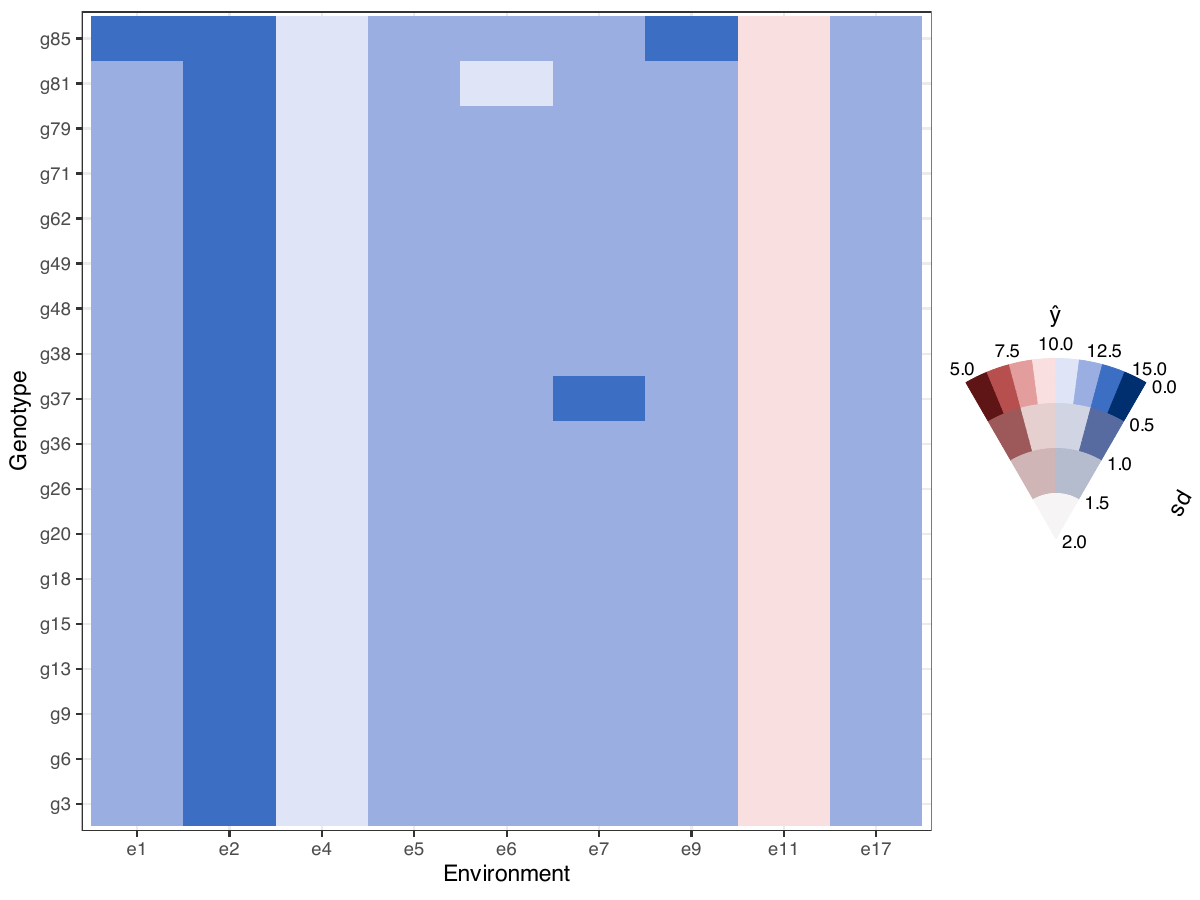}
}
\hfill
\subfloat[2015]{\label{y2015}
\centering
\includegraphics[width=0.48\columnwidth]{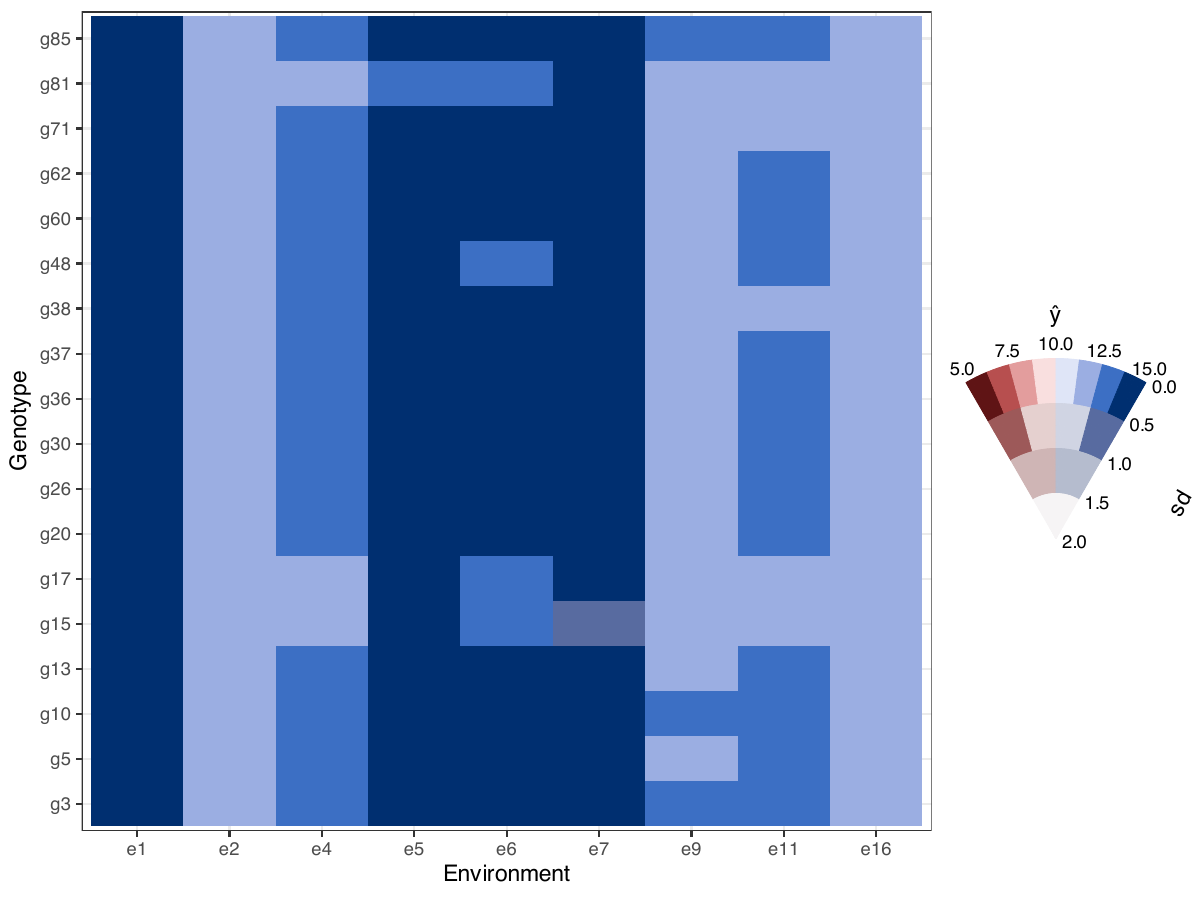}
}
\end{figure}

\begin{figure}[H] \ContinuedFloat
\subfloat[2016]{\label{y2016}
\centering
\includegraphics[width=0.48\linewidth]{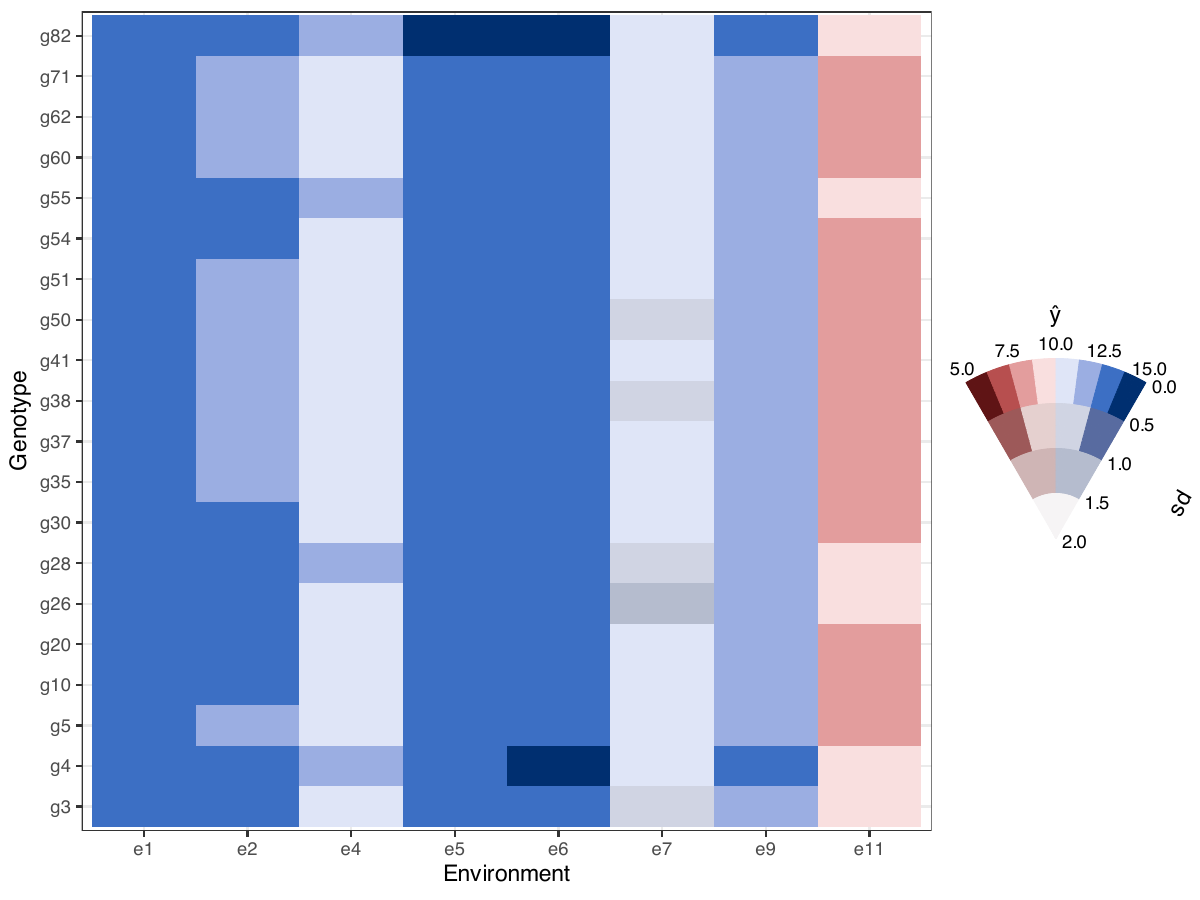}
}
\hfill
\subfloat[2017]{\label{y2017}
\centering
\includegraphics[width=0.48\linewidth]{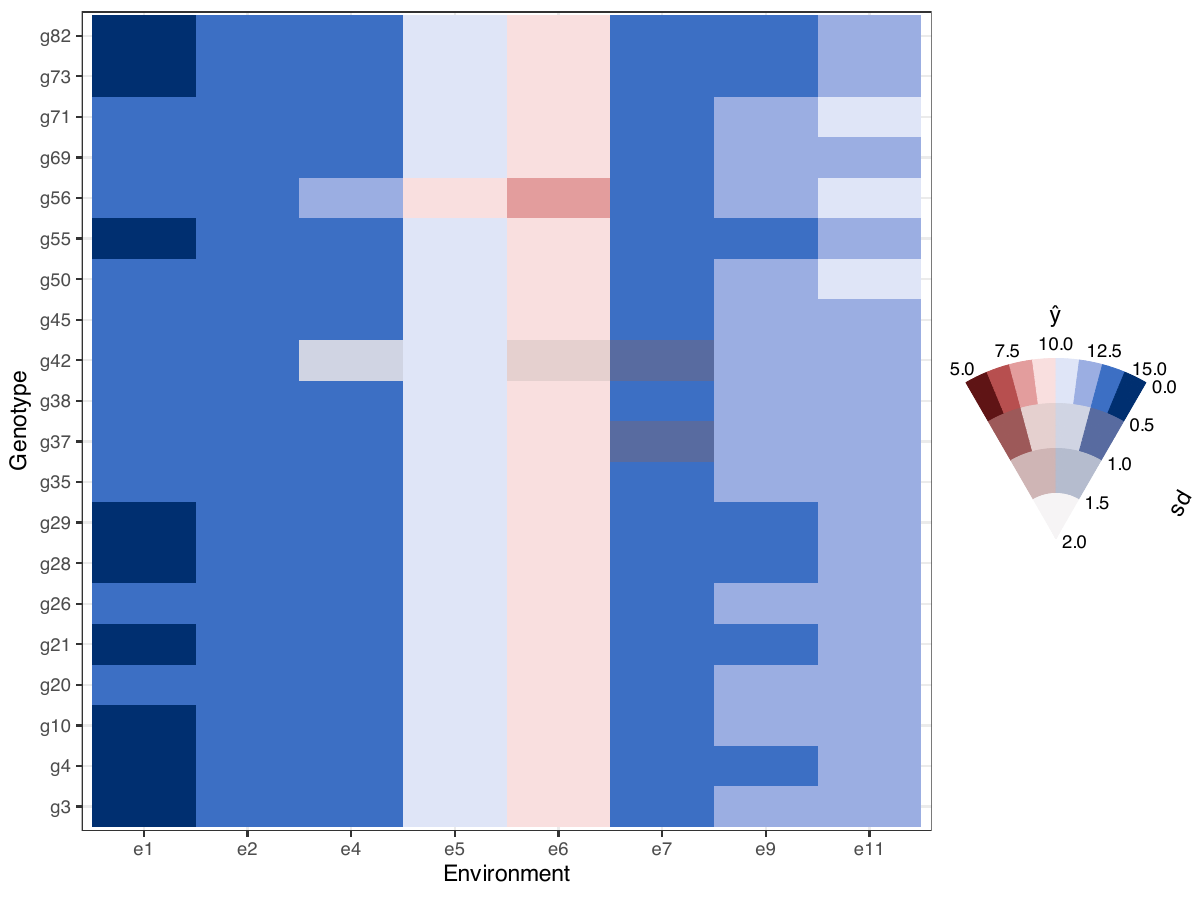}
}\\
\subfloat[2018]{\label{y2018}
\centering
\includegraphics[width=0.48\linewidth]{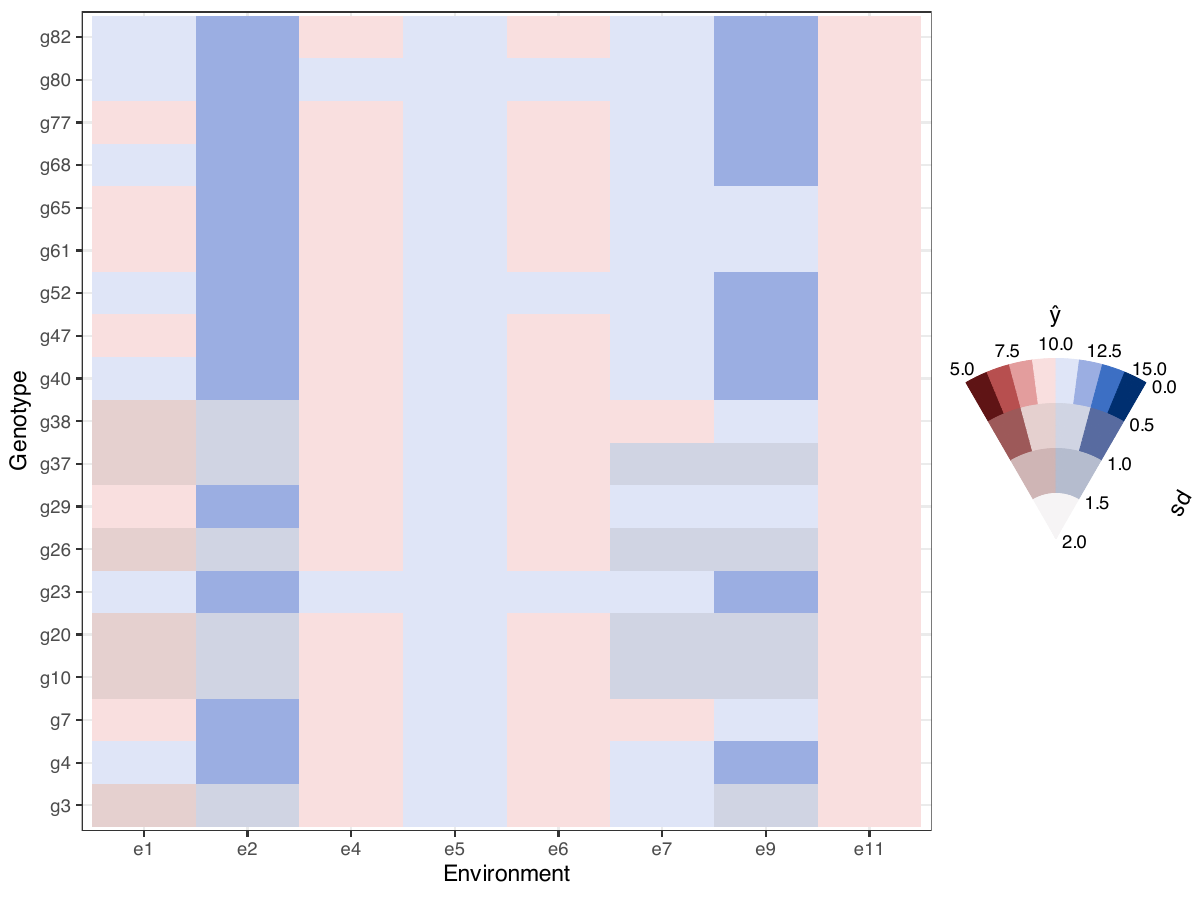}
}\hfill
\subfloat[2019]{\label{y2019}
\centering
\includegraphics[width=0.48\linewidth]{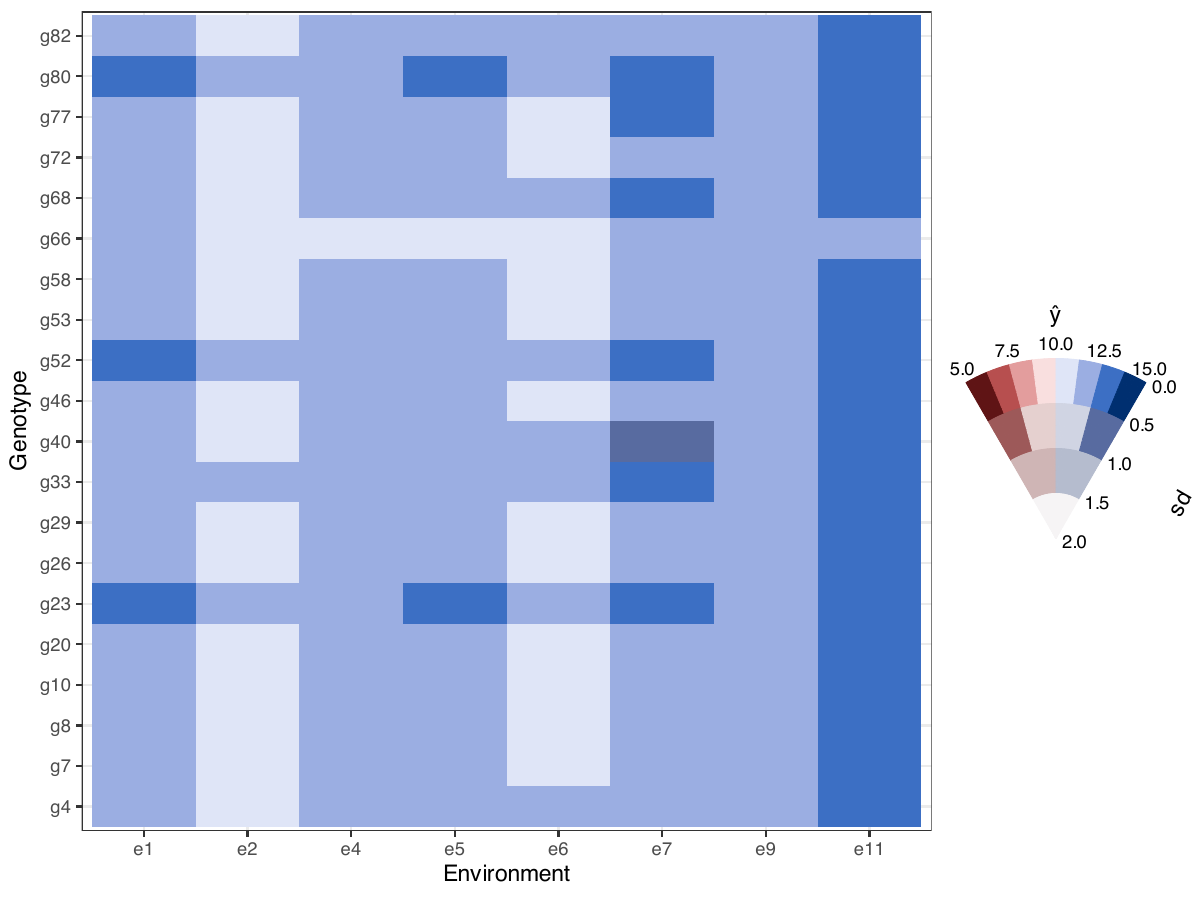}
}
\caption{Predicted yields from the BAMMIT model across the Irish dataset, with consistent ordering and colour scales for ease of comparison.}
\label{fig:vsup_all}
\end{figure}

\bibliographystyle{biometrical}
\bibliography{bib.bib}


\end{document}